\documentclass[journal]{IEEEtran}

\usepackage{hyperref}       
\usepackage{url}            
\usepackage{booktabs}       
\usepackage{amsfonts}       
\usepackage{nicefrac}       
\usepackage{microtype}      
\usepackage{xcolor}         
\usepackage{graphicx}
\usepackage{makecell}
\usepackage{arydshln, colortbl, diagbox, lettrine, marginnote, picinpar, sidecap}
\usepackage{amsmath}
\usepackage{wrapfig}
\usepackage{multirow}
\usepackage{cite}
\usepackage{float}

\usepackage{times}
\usepackage{latexsym}
\usepackage{amssymb,bbm}
\usepackage{bm}
\usepackage{pgfplots}
\usepackage{latexsym}
\usepackage{subfigure}
\usepackage{color}
\usepackage{verbatim}

\makeatletter
\def\ps@IEEEtitlepagestyle{%
  \def\@oddfoot{\mycopyrightnotice}%
  \def\@oddhead{\hbox{}\@IEEEheaderstyle\leftmark\hfil\thepage}\relax
  \def\@evenhead{\@IEEEheaderstyle\thepage\hfil\leftmark\hbox{}}\relax
  \def\@evenfoot{}%
}
\def\mycopyrightnotice{%
  \begin{minipage}{\textwidth}
  \centering \scriptsize
  Copyright~\copyright~2022 IEEE. Personal use of this material is permitted. Permission from IEEE must be obtained for all other uses, in any current or future media, including\\reprinting/republishing this material for advertising or promotional purposes, creating new collective works, for resale or redistribution to servers or lists, or reuse of any copyrighted component of this work in other works by sending a request to pubs-permissions@ieee.org.
  \end{minipage}
}
\makeatother

\begin{document}

\title{Rethinking Textual Adversarial Defense for Pre-trained Language Models}

\author{Jiayi~Wang,
        Rongzhou~Bao,
        Zhuosheng~Zhang,
        Hai~Zhao
\thanks{This paper was partially supported by Key Projects of National Natural Science Foundation of China (U1836222 and 61733011).}
\thanks{Jiayi Wang, Zhuosheng Zhang, and Hai Zhao are with the Department of Computer Science and Engineering, Shanghai Jiao Tong University, and also with Key Laboratory of Shanghai Education Commission for Intelligent Interaction and Cognitive Engineering, Shanghai Jiao Tong University, and also with MoE Key Lab of Artificial Intelligence, AI Institute, Shanghai Jiao Tong University. Rongzhou Bao is with Ant Group. Part work was done when Jiayi Wang was in Ant Group and Rongzhou Bao was in Shanghai Jiao Tong University. (E-mail: wangjiayi\_102\_23@sjtu.edu.cn; rongzhou.brz@antgroup.com; zhangzs@sjtu.edu.cn; zhaohai@cs.sjtu.edu.cn).}
\thanks{Corresponding author: Hai Zhao}
\thanks{Part of this study has been accepted as ``Defending Pre-trained Language Models from Adversarial Word Substitution Without Performance Sacrifice'' \cite{lastwork} in the Findings of the Association for Computational Linguistics: ACL-IJCNLP 2021. This paper extends the previous word-level defense framework to a universal defense framework. We further conduct experiments to investigate the properties of adversarial examples faced with detection and randomization.}
}

\markboth{IEEE/ACM TRANSACTIONS ON AUDIO, SPEECH, AND LANGUAGE PROCESSING}%
{}

\maketitle

\begin{abstract}
Although pre-trained language models (PrLMs) have achieved significant success, recent studies demonstrate that PrLMs are vulnerable to adversarial attacks. By generating adversarial examples with slight perturbations on different levels (sentence / word / character), adversarial attacks can fool PrLMs to generate incorrect predictions, which questions the robustness of PrLMs. However, we find that most existing textual adversarial examples are unnatural, which can be easily distinguished by both human and machine. Based on a general anomaly detector, we propose a novel metric (Degree of Anomaly) as a constraint to enable current adversarial attack approaches to generate more natural and imperceptible adversarial examples. Under this new constraint, the success rate of existing attacks drastically decreases, which reveals that the robustness of PrLMs is not as fragile as they claimed. In addition, we find that four types of randomization can invalidate a large portion of textual adversarial examples. Based on anomaly detector and randomization, we design a universal defense framework, which is among the first to perform textual adversarial defense without knowing the specific attack. Empirical results show that our universal defense framework achieves comparable or even higher after-attack accuracy with other specific defenses, while preserving higher original accuracy at the same time. Our work discloses the essence of textual adversarial attacks, and indicates that (\romannumeral1) further works of adversarial attacks should focus more on how to overcome the detection and resist the randomization, otherwise their adversarial examples would be easily detected and invalidated; and (\romannumeral2) compared with the unnatural and perceptible adversarial examples, it is those undetectable adversarial examples that pose real risks for PrLMs and require more attention for future robustness-enhancing strategies.
\end{abstract}


\IEEEpeerreviewmaketitle

\section{Introduction}

\IEEEPARstart{W}{ith} rapid development and significant success of deep neural networks (DNNs), growing concerns about the robustness of DNNs have been raised \cite{taslp1, taslp2}. Recent works demonstrate that DNNs are vulnerable to adversarial examples, which fool models to generate incorrect predictions by introducing intentional perturbations to original examples \cite{goodfellow2015explaining,new1,new2,new3,new4,new5}. Adversarial examples were first proposed in the task of image classification \cite{firstcv} in the computer vision (CV) community and received considerable attention in the last five years. The properties of image adversarial examples have been well explored and analyzed. The fact that such attacks can be detected \cite{detectionreview} and be invalidated by randomization \cite{randomization1, randomization2} has triggered many valuable following studies \cite{cannotdetect}, including designs of stronger adaptive attacks \cite{FalseSecurity} and better understanding of properties of adversarial examples \cite{DBAnalysis}. However, compared to their counterparts in computer vision, adversarial examples in the natural language processing (NLP) domain are very different due to the discrete nature of languages, and properties of textual adversarial examples have not been well studied. Thus, in this work, comprehensive experiments are conducted to explore whether textual adversarial examples possess similar properties compared to their counterparts in computer vision. For the targeted tasks, we focus on text classification and natural language inference, which are the most important and commonly investigated tasks in the adversarial attack domain. As for the targeted DNNs, we focus on pre-trained language models (PrLMs), which are widely used as an essential component for NLP systems. In this way, our research would have a broader impact on the NLP community.
 
We firstly investigate whether textual adversarial examples generated by various attack methods, ranging from character, word to sentence levels, can be detected. We conduct experiments under two scenarios: (\romannumeral1) with a specific anomaly detector (a detector targeting at a specific attack) and (\romannumeral2) with a general anomaly detector (a detector expected to detect unknown attacks of any type). We find that both specific and general anomaly detectors can easily identify adversarial examples of all levels. This indicates that although existing attacks have tried to impose restrictions on several metrics to make adversarial examples imperceptible, most adversarial examples are still unnatural and therefore can be detected and blocked. 

As the general anomaly detector is able to detect adversarial examples generated by unknown attack models, we can conclude that despite the differences in attack algorithms, the adversarial examples share some common features, which can be learned by the general anomaly detector. Based on this finding, we leverage the output probability of the general anomaly detector (Degree of Anomaly) as a perturbation constraint to ensure the imperceptibility of adversarial examples. We apply this novel constraint to state-of-the-art attack approaches, and the new attack success rate decreases sharply for attacks at all levels. The sharp decrease indicates that the PrLMs are more robust than previous attack methods claimed, as long as we set up strict but reasonable perturbation constraints.

Furthermore, we find that several randomization processes can invalidate most of the textual adversarial examples. Randomization is a well-explored defense strategy in computer vision \cite{randomization1, randomization2}. However, randomization in the NLP domain is quite different due to the discrete nature of texts. To our best knowledge, random synonym substitution is the only randomization strategy investigated in textual adversarial defenses \cite{safer, RSE}, which is used to evaluate certified robustness of a model. 

Inspired by the common practice in data augmentation \cite{EDA,DataAugmentation2}, besides random synonym substitution, we propose three novel randomization processes to defend textual attacks: (\romannumeral1) random adverb insertion, (\romannumeral2) random MLM (masked language model) prediction and (\romannumeral3) back translation. Empirical results demonstrate that over 50\% of adversarial examples can be invalidated by applying only once the randomization, which implies the vulnerability of adversarial examples to randomization. Based on this finding, we adopt randomization as a defense module to defend against dynamic attacks. In order to perplex the attack model, we apply randomization to all intermediate adversarial examples generated in each attack iteration. Experimental results demonstrate that such randomization significantly improves the prediction accuracy under adversarial attacks.

In the computer vision domain, there exist adaptive attacks bypassing randomization by adopting Expectation Over Transformation (EOT) \cite{EOT, FalseSecurity} to model randomization within the optimization procedure. Borrowing the idea of EOT, we design experiments of adaptive attacks to try bypassing textual randomization. However, the attack success rate has little increase. We analyse the reasons for the different effectiveness of EOT in image and text domains from three aspects: (\romannumeral1) search space, (\romannumeral2) optimization process and (\romannumeral3) role of randomization in Section \ref{sec:adaptive}.

Based on the impressive performance of the anomaly detector and randomization on recognizing and defending adversarial examples, we integrate these two methods into a universal defense framework to help PrLMs defend against attacks at all levels. While previous defense methods all require partial \cite{RobustnessWS} or full \cite{RabbitHole} knowledge of targeted attack patterns, our universal defense framework is the first one that is capable of resisting unknown types of attacks, and therefore can be easily applied to all kinds of attacks without extra training. By leveraging this universal defense framework, all state-of-the-art textual attacks examined in this study can be largely defended. Compared with other existing defenses targeting specific attacks, our universal defense framework can achieve comparable or even higher after-attack accuracy while better preserving original precision at the same time. This result indicates that further works of adversarial attacks should focus more on how to overcome two issues: (\romannumeral1) how to avoid being distinguished from normal examples and (\romannumeral2) how to avoid being invalidated by randomization. Otherwise they would be easily detected and defended by our defense framework.

Our contributions are summarized as follows:

1) We demonstrate that existing textual adversarial examples can be easily detected. By leveraging the output probability of a general anomaly detector, we propose anomaly score as a novel perturbation constraint to help ensure the imperceptibility of adversarial examples. Empirical results show that effectiveness of different attacks sharply decreases under the new constraint, revealing that PrLMs are more robust than previous works claimed.

2) We find that textual adversarial examples are unstable to four types of randomization. Experiments demonstrate that randomization during attack can successfully defend against state-of-the-art adversarial attacks of all types. Furthermore, we design adaptive attacks based on the concept of EOT in CV to bypass textual randomization. Experiments of adaptive attacks further prove the robustness and effectiveness of randomization as a defense method.

3) We combine detector and randomization to form a universal defense framework that has three advantages: (\romannumeral1) it is among the first defenses that can defend against all kinds of adversarial attacks without knowing the specific attack algorithms; (\romannumeral2) it has comparable or even higher after-attack accuracy with other defenses targeting specific attack method; (\romannumeral3) it preserves higher original accuracy of the model, which is important in realistic application; (\romannumeral4) it can be modified flexibly by tuning the implementation of randomization.

\section{Terminology}
\begin{itemize}
    \item \textit{Perturbation.} Perturbations are intently designed small noises that are added to original examples to fool the victim model.
    \item \textit{Adversarial Example.} Adversarial examples are created by an attack model via adding small perturbations in original examples such that the victim model makes wrong predictions. At the same time, adversarial examples should be imperceptible to humans, which means (\romannumeral1) humans cannot distinguish adversarial examples from original examples and (\romannumeral2) humans should still make correct predictions on adversarial examples.
    \item \textit{Robustness.} A model is robust if it can make correct predictions when faced with imperceptible perturbations. Adversarial defenses are designed to enhance robustness of models.
    \item \textit{Attack Model.} Attack model refers to the model that generates adversarial examples.
    \item \textit{Victim Model.} Victim model is the model under attack. The victim models investigated in this paper are PrLMs.
\end{itemize}

\section{Background}

\subsection{Textual Adversarial Attacks}

\subsubsection{Problem Formulation}
Textual adversarial attacks generate adversarial examples against a victim model $F$, which we assume is a text classifier based on PrLM. Given a dataset of $N$ input sequences $\mathcal{X} =\{X_1, \dots, X_N\}$ and their labels $\mathcal{Y} =\{Y_1,  \dots, Y_N\}$, the victim model $F: \mathcal{X} \rightarrow \mathcal{Y}$ maps the input space $\mathcal{X}$ to the label space $\mathcal{Y}$. For an input sequence $X$ in $\mathcal{X}$, its corresponding adversarial example $X_{adv}$ should satisfy:
\begin{equation}
F(X_{adv}) \neq F(X),\quad and \quad d(X_{adv}-X) \leq \sigma,
\end{equation}
where $d()$ measures the perceptual difference between $X_{adv}$ and $X$, and $\sigma$ is a threshold to limit the size of perturbations.

\subsubsection{From Image Attacks to Textual Attacks}
Adversarial attacks originated from the computer vision community and a diversity of attack algorithms and defenses have been proposed along this research direction in the last five years. As \cite{cvsurvey} summarizes, attack and defense techniques for images can be classified into first-generation techniques (core algorithms focusing on classification problem) before 2018 and second-generation techniques (methods that further refine and adapt core algorithms on various vision tasks) until now. First-generation techniques including Fast Gradient Sign Method (FGSM) \cite{AdvTraining1}, Jacobian Saliency Map Adversary (JSMA) \cite{JSMA},  C\&W Attack \cite{CW} have inspired many earliest works of white-box textual attacks \cite{Textfool, JSMA-based, CW-based} around the year of 2018.

However, although ideas can be borrowed, we cannot directly apply the approaches of image attacks to textual attacks because of two traits of natural language:
\begin{itemize}
    \item \textit{Discrete Inputs.} Image inputs are continuous pixels, while textual inputs are discrete language symbols. Although textual sequence can be mapped to a vector in a continuous high-dimensional space, adding continuous noise directly to vectors can lead to out-of-vocabulary words or meaningless phrases. 
    \item \textit{Semantics and Grammar.} In computer vision, small changes to image pixels are usually imperceptible to humans and will not change the meaning of the image. In NLP, however, any small changes have the potential to alter the semantic meaning of the sentence; or break the grammar rules; or influence the fluency of language. For example, inserting a negation word like ``not" before ``happy" will change a sentence of positive sentiment into a sentence of negative sentiment. In addition, the change of words or characters can also be easily perceived by humans.
\end{itemize}

Due to these traits of natural language, textual attack models either carefully adjust the methods from image attacks with additional constraints, or propose novel methods. The latter direction is more popular in recent years \cite{TextFooler, BertAttack}.

\subsubsection{Perturbation Constraint \label{sec:perturbation constraint}} 
In the domain of computer vision, input images are represented by continuous vectors and adversarial examples are designed by adding perturbations in pixel values. Thus the perceptual difference function $d()$ can be easily defined by Euclidean distance $d(X_{adv}-X) = \left\| X_{adv}-X\right\|_2$ or the Chebyshev distance $d(X_{adv}-X) = \left\| X_{adv}-X\right\|_\infty$. However, in the domain of NLP, changing a character or a word in a sentence can be easily perceived by humans, or can largely alter the semantic meaning of the original sentence. Therefore, the definition of $d()$ for textual perturbations has to take semantic similarity and syntactic fluency into consideration. Previous works propose several heuristic metrics to define $d()$:

\begin{itemize}
    \item \textit{Semantic similarity.}  Many existing attack works \cite{TextFooler} \cite{BertAttack} use Universal Sentence Encoder (USE) \cite{USE} to encode original sentence and adversarial sentence into high dimensional vectors and use their cosine similarity as an approximation of semantic similarity. Given two vectors $x$ and $y$ of $n$ dimensions, their cosine similarity is: 
    \begin{equation}
    CosSim(x,y) = \frac{x \cdot y}{\vert\vert x \vert\vert \cdot \vert\vert y \vert\vert} 
    \end{equation}

    \item \textit{Perturbation Rate.} Perturbation rate is often used in word-level attacks \cite{TextFooler} \cite{BertAttack} to indicate the rate between number of modified words and total words. 
    
    \item \textit{Number of increased grammar errors.} It is the absolute number of increased grammatical errors in the successful adversarial example, compared to the original text.  This metric is used in \cite{HardLabel}, \cite{clare} and is calculated using LanguageTool \cite{languagetool}.
    
    \item \textit{Levenshtein distance.} Levenshtein distance is a kind of edit distance which refers to the number of
    editing operations to convert one string to another. It is often used in character-level attacks \cite{deepwordbug} to measure the similarity between two strings.

    \item \textit{Perplexity.} Perplexity is a metric to evaluate the fluency of adversarial examples \cite{perplexity, combinatorial, clare}. The perplexity is calculated using small sized GPT-2 with a 50K-sized vocabulary \cite{gpt2}.

    \item \textit{Jaccard similarity coefficient.} The Jaccard similarity coefficient is used to measure the similarity between finite sample sets using intersection and union of the sets. For two given sequences $A$ and $B$, their Jaccard similarity coefficient $J(A, B)$ is calculated by
    \begin{equation}
    J(A, B) = \frac{|A \cap B|}{|A \cup B| }.
    \end{equation}
    where $|A \cap B|$ denotes the number of words appearing in both sequences, $|A \cup B|$ refers to the number of total words without duplication, thus $0 \leq J(A, B) \leq 1$. The closer the value of $J(A, B)$ is to 1, the more similar $A$ and $B$ are. 
    
    \item \textit{Human evaluation.} Besides above automatic metrics, human evaluation is also used in some attack works \cite{BertAttack, clare}.  They ask crowd-sourced judges to score grammar correctness and semantic similarity, and make predictions on adversarial examples. This metric directly applies the definition of adversarial examples, but is expensive and hard to replicate. 
    
\end{itemize}

\subsubsection{Effectiveness Evaluation}
The effectiveness of textual adversarial attacks is evaluated from three aspects: (\romannumeral1) decrease of accuracy of victim model, (\romannumeral2) quality of generated adversarial examples and (\romannumeral3) algorithm efficiency. The commonly used measurements are: 
\begin{itemize}
    \item \textit{After-attack accuracy.} After-attack accuracy (or adversarial accuracy) is the accuracy of the victim model against the adversarial examples crafted from the test examples. Lower after-attack accuracy implies stronger attack.
    \item \textit{Attack success rate.} Attack success rate is the ratio of number of wrong predictions to the total number of adversarial examples.
    \item \textit{Perturbation constraints.} The perturbation constraints like semantic similarity, perturbation ratio and number of grammar errors are also used as metrics to measure the quality of generated adversarial examples. 
    \item \textit{Query number.} For black-box attacks \cite{TextFooler, BertAttack}, queries of the target model are the only accessible information. Too many queries are time-consuming and not feasible in reality. Thus average query number can reflect the efficiency of black-box attacks.
    
\end{itemize}

\subsubsection{Categories of Textual Adversarial Attacks}
\begin{itemize}
    \item \textit{Model knowledge.} According to their knowledge of victim models, textual attacks can be classified into white-box attacks \cite{whitebox} (require full information of the victim model) and black-box attacks \cite{BertAttack} (only require the output of the victim model). Black-box attacks can be further categorized into score-based attacks \cite{TextFooler} (require the victim model's prediction score as output) and decision-based attacks \cite{HardLabel} (only require the victim model's prediction label as output).
    \item \textit{Language granularity.} According to language granularity of perturbations, textual attacks can be classified into character-level, word-level and sentence-level attacks. Character-level attacks \cite{deepwordbug} perturb original sentences by manipulating characters in several words, such as replacing, inserting or deleting a character. Word-level attacks substitute several words by their synonyms in either a heuristic \cite{TextFooler} or a contextualized \cite{BertAttack,garg2020bae} way to fool the model. Sentence-level attacks generate adversarial examples by paraphrasing the original sentence \cite{scpn} or using a generative adversarial network (GAN) \cite{gan}. 
    \item \textit{Target.} According to the objective of attacks, textual attacks can be divided into targeted and non-targeted attacks. For targeted attacks, each generated adversarial example should be classified into a specified target class. For non-targeted attacks, the attack model is only required to fool the model to make incorrect predictions. For binary classification, non-targeted attacks and targeted attacks are the same. In this paper, all investigated attacks are non-targeted attacks.
    
\end{itemize}

\subsection{Textual Adversarial Defenses}

\subsubsection{Problem Definition}
The aim of adversarial defenses is to learn a
model that can achieve high test accuracy on
both benign and adversarial examples. Adversarial defenses should not only defend against fixed adversarial examples, but also defend against iterative attacks. In the scenario of defense, attacks are assumed to have access to the defense model and can attack the defense model iteratively to generate adversarial examples.

\subsubsection{Effectiveness Evaluation}
The effectiveness of a defense model is evaluated from two aspects: (\romannumeral1) improvement of robustness and (\romannumeral2) preservation of precision. The corresponding metrics are:
\begin{itemize}
    \item \textit{After-attack accuracy.} Higher the after-attack accuracy is, more robust the defense model is.
    \item \textit{Original accuracy.} It is important that the defense model preserves the original accuracy on benign examples. Since in reality, it is not reasonable to sacrifice too much precision for potential security risks.
\end{itemize}

\subsubsection{Categories of Textual Adversarial Defenses}
The research directions of textual adversarial defenses can be classified into three broad categories:
\begin{itemize}
    \item \textit{Adversarial training.} As common practices, by leveraging adversarial examples to conduct data augmentation, adversarial training \cite{AdvTraining1, HardLabel} is widely adopted to increase the robustness of victim models.  However, such adversarial training techniques are subject to the limited number of adversarial examples. Empirical results \cite{TextFooler, HardLabel} indicate that the increase of robustness brought by adversarial training alone is quite limited.
    \item \textit{Adversarial restoration.} The idea of adversarial restoration is to detect and then reconstruct perturbed tokens. For character-level attacks, ScRNN \cite{CombatMisspellings} adopts an RNN semi-character architecture to identify and restore words corrupted by character manipulations. For word-level attacks, DISP \cite{DISP} restores adversarial examples by using a perturbation discriminator and an embedding estimator, both of which are based on BERT \cite{bert}. FGWS \cite{FGWS} exploits the frequency properties of adversarial word substitutions to identify perturbed words.
    \item \textit{Certified robustness.} Interval Bound Propagation (IBP) \cite{IBP} has been proposed to consider theoretically the worst-case perturbation in order to certify the robustness of models \cite{CertifiedRobustness, CertifiedRobustness2}. SAFER \cite{safer} proposes another certified robust method based on randomized smoothing technique. However, certified robustness requires a strong constraint on attack space and is hard to scale to large datasets and neural networks due to the high complexity. In addition, it does harm to the model's accuracy on clean examples due to the looser upper bounds.
\end{itemize}

To the best of our knowledge, most defense methods either only target a certain type of attack or require the knowledge of targeted attack, which limits their effectiveness in real application scenarios.  

\subsection{Anomaly Detection and Randomization}
In this paper, we mainly investigate two properties of textual adversarial examples: their detectability and their vulnerability to randomization. Part of our method is based on our previous study \cite{lastwork}. We extend the idea of anomaly detection and randomization in the previous work to construct a universal defense framework. In this subsection, we summarize existing works on anomaly detection and randomization for both image and text domain, then analyse the difference between our work and these previous works.

\subsubsection{Anomaly Detection}
\begin{itemize}
    \item \textit{In CV.} Anomaly detection is a well-explored research direction in CV \cite{cv-survey}. We introduce some of the state-of-the-art detection methods proposed in the last three years. LiBRe \cite{cv-detection2} (Lightweight Bayesian Refinement) replaces the last few layers of victim model by a lightweight Bayesian neural network (BNN) to distinguish adversarial perturbations. In \cite{cv-detection3}, the authors leveraged a mechanism to trace the activation paths of benign and adversarial images and conducted detection based on the difference in the features of these paths. In \cite{cv-detection1}, the authors proposed a mechanism of class-conditional reconstruction of images to detect adversarial examples.  

    \item \textit{In NLP.} Compared with CV, there are few works on anomaly detection in NLP and we list all the works we know here. For character-level attacks, robust word recognition \cite{CombatMisspellings} places a word recognition model in front of the downstream classifier to detect adversarial spelling mistakes. For word-level attacks,  DISP \cite{DISP}, a framework called learning to discriminate perturbations, is proposed to identify malicious perturbations. More recently, FGWS \cite{FGWS} leverages frequency-guided word substitutions (FGWS), exploiting the frequency properties of adversarial word substitutions for the detection of adversarial examples. For sentence-level attacks, Darcy \cite{darcy} is a honeypot-based detection method against the universal trigger attack method. All of these detection methods target one-level attacks and cannot generalize to adversarial examples of other levels.
    
    \item \textit{In this paper.} We adopt a very naive but effective detector: a binary classifier based on PrLM. Different from existing works of anomaly detection in NLP which target at a specific type of attack, we demonstrate that all kinds of adversarial examples can be detected with a high F1 score by this simple detector. The emphasis of our work is not to design a complicated detector, but to investigate the detectability of all types of adversarial examples, and further utilize the common features of different adversarial examples to propose a novel metric filtering unnatural perturbations. In addition, several directions for further enhancing the detector are discussed in Section \ref{sec:discussion}.  
    
\end{itemize}

\subsubsection{Randomization}

\begin{itemize}
    \item \textit{In CV.} Since transformations on images are trivial and continuous, there have been many defense works based on image transformation to make adversarial examples benign for the victim model \cite{cv-survey}. The transformation-based defense methods include: JPEG-based compression \cite{cv-rand1}, \cite{cv-rand2}, stochastic combination of multiple input transformations \cite{cv-rand3}, set of random input transformations \cite{cv-rand4}, learning-based compression techniques \cite{cv-rand5}. 
    
    However, an adaptive attack called Expectation over Transformation (EOT) \cite{FalseSecurity} \cite{EOT} is proposed to bypass defenses based on random transformations. The idea of EOT is to model random transformations within the optimization procedure. When attacking a victim model $f(·)$ that randomly transforms the input image $x$ according to a transformation function $t(·)$ sampled from a distribution of transformations $T$, EOT optimizes the expectation over the transformation $\mathbb{E}_{t \sim T} f(t(x))$. As image input is continuous, the optimization problem can be solved using gradient descent. Noting that $\nabla \mathbb{E}_{t \sim T} f(t(x)) = \mathbb{E}_{t \sim T} \nabla f(t(x))$, EOT differentiates through the classifier and transformation, and approximates the expectation at each gradient descent step.
    
    \item \textit{In NLP.} The randomization-based defenses in NLP have not been well investigated. To our best knowledge, random synonym substitution is the only randomization strategy proposed by previous works targeting word-level attacks. We introduce the only two existing textual defense works related to randomization here. RSE \cite{RSE} (Random Substitution Encoding), introduces a random substitution encoder into the training process of the original model. SAFER \cite{safer} is a certified robust method based on a randomized smoothing technique, which constructs a stochastic ensemble by applying random synonym substitutions on the input sentences, and leverages the statistical properties of the ensemble to provably certify the robustness. 
    
    \item \textit{In this paper.} We investigate random synonym substitution and three new randomization strategies: random adverb insertion, random MLM suggestion and back translation. Different from the two previous works which adopt randomness in encoder or as a smoothing technique to certify the robustness, we directly adopt randomization as a defense method. We also adapt the idea of Expectation Over Transformations proposed in CV to design adaptive attacks against randomization in the textual domain. However, these adaptive attacks are not as efficient as their counterparts in CV. We analyse the reasons for this difference in Section \ref{sec:adaptive}.
    
\end{itemize}

\section{Methods}
Our methods contain three main parts: (\romannumeral1) Anomaly Detection; (\romannumeral2) Randomization; (\romannumeral3) Universal Defense Framework.

Before introducing the methods, we would like to explain why the combination of anomaly detection and randomization is a promising method to defend against textual attacks. Firstly, existing textual defenses implement the same mechanisms for normal and adversarial examples, which leads to a decrease in the model’s performance on normal examples. Therefore it is natural to consider if we could distinguish between normal and adversarial examples and only conduct defense on adversarial examples. The feasibility of this idea is based on a discriminator of high accuracy. Since we find that most adversarial examples are unnatural and easy to distinguish, this direction becomes promising. 

Secondly, many image defenses are based on random transformations, but surprisingly this direction is seldom investigated in textual defenses. So we are curious about the utility of randomization in textual defense and investigate several randomization settings. We consider randomization as a promising defense direction for two reasons: (\romannumeral1) an adversarial example can be seen as an especially perturbed example that is located very near to the decision boundary but at the wrong subarea. It is hopeful that randomization can pull it back to the correct subarea. (\romannumeral2) The attacks we investigate use the output score or decision of the victim model to direct the generation of adversarial examples. Randomization can perplex the feedback score got by the attacks.

\begin{figure*}
  \centering
  \includegraphics[width=0.88\textwidth]{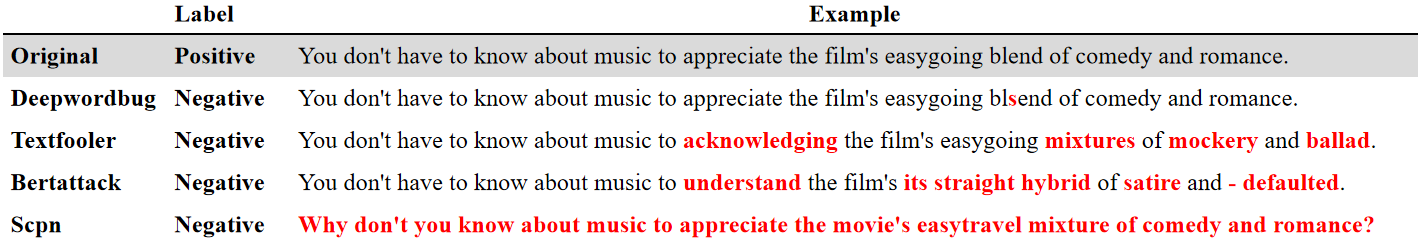}
  \caption{Examples of adversarial examples generated by different levels of attacks.}
  \label{fig:attack}
\end{figure*}

\subsection{Anomaly Detection}
As shown in Figure \ref{fig:attack}, despite the existence of perturbation constraints, some adversarial examples generated by different attacks are still not natural enough and thus can be perceived by humans. We wonder if machine can also perceive the anomaly features in adversarial examples. Thus we train an anomaly detector to distinguish adversarial examples from normal examples. For an input sequence $X=\{x_1, \dots, x_n\}$, $X$ is firstly separated into sub-word tokens (with a special token \texttt{[CLS]} added at the beginning), and converted into embeddings $E=\{e_1, \dots, e_m\}$. A PrLM then captures the contextual information by attention mechanism and generates a sequence of contextual embeddings $\{h_0, h_1, h_2, ..., h_m\}$, in which $h_0 \in \mathbb{R} ^H$ is the contextual representation of special token \texttt{[CLS]}. For text classification tasks, $h_0$ is used as the aggregate sequence representation. The anomaly detector leverages $h_0$ to predict the probability that $X$ is labeled as class $\hat{y_d}$ (if $X$ is adversarial example, $\hat{y_d} = 1$; if $X$ is normal example, $\hat{y_d} = 0$) by a logistic regression with softmax:
\begin{equation}
y_d = Prob(\hat{y_d}|X) = {\rm softmax}(W_d(dropout(h_0))+b_d).
\end{equation}

The loss function is:
\begin{equation}
loss_d = -[y_d*{\rm log}
\hat{y_d}+(1-y_d)*{\rm log}
(1-\hat{y_d})].
\end{equation}

\subsubsection{Specific Anomaly Detector} 
Firstly, we train a specific anomaly detector for each investigated attack to distinguish its adversarial examples from normal examples. For each dataset, the training data consist of normal examples from its train set (labeled as 0) and their corresponding adversarial examples (labeled as 1) generated by the targeted attack. 

\subsubsection{General Anomaly Detector} 
A limitation of specific anomaly detector is that we need to know in advance the attack method to train the corresponding specific anomaly detector. But is the knowledge of specified attack methods really necessary to perform anomaly detection? We further investigate if we can train a general anomaly detector which can distinguish adversarial examples of all types, without knowing in advance the attack method. 

To train this general anomaly detector, we use all the original examples (labeled as 0) and adversarial examples (labeled as 1) from train sets of all datasets and all attack methods used in the training stage of the specific anomaly detectors.

\subsubsection{Quantifying Anomaly} We define a new metric: \textbf{Degree of Anomaly} to quantify the anomaly of a sentence. Given a sentence $X$, we leverage the probability that $X$ is adversarial example predicted by general anomaly detector as the Degree of Anomaly of $X$:  
\begin{equation}
Degree(X) = Prob(\hat{y_d} = 1|X).
\end{equation}

\subsubsection{Attack under Anomaly Constraint} In order to enable attacks to generate more natural and undetectable adversarial examples, we use degree of anomaly as a new metric to constrain the size of perturbations. The attack problem formulation now becomes:
\begin{equation}
\begin{aligned}
&F(X_{adv}) \neq F(X)\\
& \begin{array}{r@{\quad}l}
s.t.& d(X_{adv}, X) < \sigma, \\
& Degree(X_{adv}) < 0.5, \\
\end{array}
\end{aligned}
\end{equation}
where $d()$ measures the perceptual difference between $X_{adv}$ and $X$. Each attack has its own definition of $d()$ and its corresponding threshold $\sigma$. For each attack, We add on a new constraint that the degree of anomaly of $X_{adv}$ should be smaller than 0.5.

\subsection{Randomization}

\begin{figure*}
  \centering
  \includegraphics[width=0.88\textwidth]{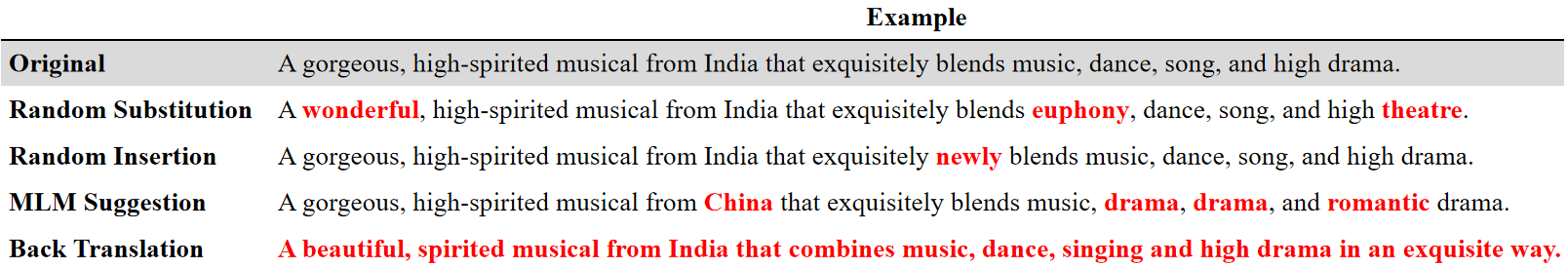}
  \caption{Examples of Randomization processes.}
  \label{fig:randomization}
\end{figure*}

Inspired by the common practice in data augmentation \footnote{\href{https://github.com/textflint}{https://github.com/textflint}}, we propose four randomization processes to mitigate adversarial effects of adversarial examples: \textbf{random synonym substitution}, \textbf{random adverb insertion}, \textbf{random MLM suggestion} and \textbf{back translation}. For each sentence, random substitution randomly selects 25\% of words and substitutes them with a random synonym chosen from their synonym set by referring to WordNet \cite{CounterFittingWordVectors}. Random adverb insertion transforms a sentence by inserting randomly chosen neutral adverbs before verbs. The list of neutral adverbs can be found here
\footnote{\href{http://textflint.oss-cn-beijing.aliyuncs.com/download/UT\_DATA/neu\_adverb\_word\_228.txt}{http://textflint.oss-cn-beijing.aliyuncs.com/download/UT\_DATA/neu\_adverb\_\\word\_228.txt}}. We only choose neutral adverbs so that the inserted adverbs will not alter the semantic meaning of the sentence. Random MLM suggestion replaces one random syntactic element in the original sentence based on the prediction given by MLMs. Back translation paraphrases a sentence by translating it into another language and translating it back. For all examples, we firstly apply a contextual spelling checker \footnote{\href{https://github.com/R1j1t/contextualSpellCheck}{https://github.com/R1j1t/contextualSpellCheck}} to restore the misspellings. The examples of the four randomization strategies are presented in Fig. \ref{fig:randomization}. 

\begin{figure*}[htb!]
  \centering
  \includegraphics[width =0.88\textwidth]{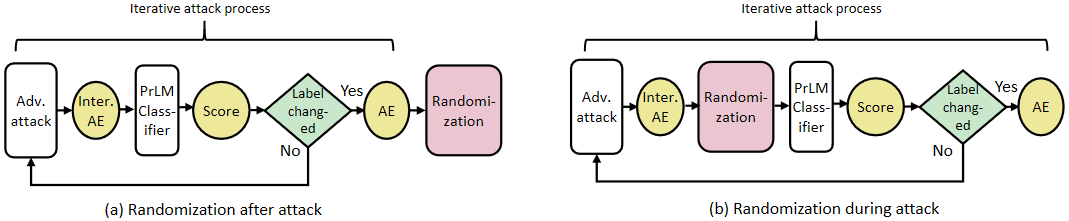}
  \caption{\label{fig:randomizationPipeline}Different positions of randomization after (a) and during (b) attack.}
\end{figure*}

We analyze the effect of different randomization processes in two settings: (a) randomization after attack and (b) randomization during attack.We firstly investigate how many adversarial examples are vulnerable to randomization. This corresponds to the setting of randomization after attack: as shown in Figure \ref{fig:randomizationPipeline} (a), we assume that adversarial examples have already been generated by the whole attack process, and we apply randomization once to these fixed adversarial examples.

Then we verify if randomization can be adopted as a defense module to defend the dynamic attack process. Thus we conduct experiments of randomization during attack. As shown in Figure \ref{fig:randomizationPipeline} (b), in each iteration, the attack process sends an intermediate adversarial example to the victim PrLM-based classifier to get its score as feedback, and we apply randomization to this intermediate adversarial example before it reaches the PrLM-based classifier. 

\subsection{Universal Defense Framework}

\label{sec:defense}
\begin{figure*}
  \centering
  \includegraphics[width =0.7\textwidth]{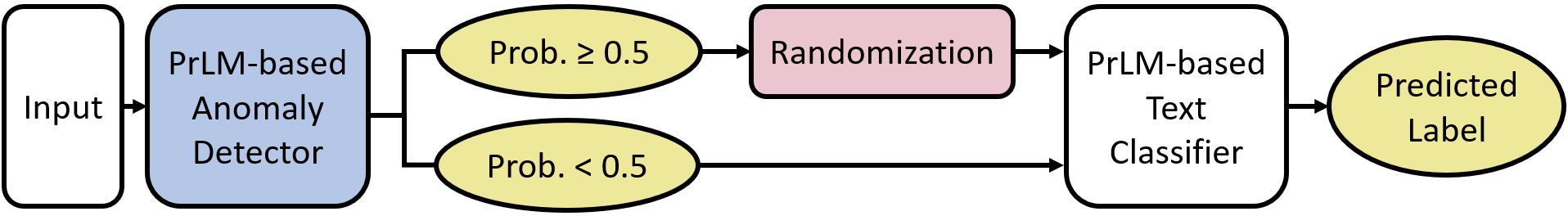}
  \caption{\label{fig:framework}Universal defense framework.}
\end{figure*}

Based on the general anomaly detector and the randomization during attack, we design a universal defense framework, which is able to defend attacks of different levels without knowing the information of coming attacks. The universal defense framework is shown in Figure \ref{fig:framework}. For each input sequence $X$, we firstly apply the general anomaly detector $F_d$ to identify whether it is abnormal ($F_d(X) = 1$) or not ($F_d(X) = 0$). If the input sequence $X$ is recognized as abnormal, a randomization process is adopted to mitigate its abnormal effect, $X'=randomized(X)$; otherwise $X$ remains unchanged, $X'=X$. Then, $X'$ is sent to the PrLM-based text classifier $F_c$ to predict its output label $F_c(X')$.  In addition, in the training stage of the PrLM-based text classifier, we augment the original training set with its randomized version, in order to make the classifier more robust to randomized benign examples.

\section{Experimental Implementation}

\subsection{Investigated Textual Attacks}
We investigate four classic textual attack methods from character level, word level to sentence level. Examples of adversarial examples generated by these four attacks are demonstrated in Figure \ref{fig:attack}.

\subsubsection{Character-level Attack}
We select DeepWordBug \cite{deepwordbug} as the representative of character-level attack. Deepwordbug applies four types of character-level modifications to a word: random character substitution, random character insertion, random character deletion and neighboring character swap. Levenshtein distance is adopted as a metric to constrain the similarity between original and adversarial sentences.

\subsubsection{Word-level Attack}
We select two classic word-level attack methods: TextFooler \cite{TextFooler} and BertAttack \cite{BertAttack}. They attack the targeted model iteratively in three steps: (\romannumeral1) sort the words in original sentence by their importance scores, (\romannumeral2) replace one word a time with a substitution word based on the importance sequence, (\romannumeral3) stop if the targeted model is fooled, otherwise continue to substitute the next word. Their difference lies in the way to find the substitution word: TextFooler forms a synonym set for each substituted word by referring to WordNet \cite{CounterFittingWordVectors} and selects the substitution from this set. While BertAttack masks the original word and uses MLM to predict the substitution word, the substitution process is context-aware and thus can generate more fluent adversarial examples.

\subsubsection{Sentence-level Attack}
We select SCPN \footnote{\href{https://github.com/thunlp/OpenAttack}{https://github.com/thunlp/OpenAttack}} \cite{scpn} to generate sentence-level adversarial examples. SCPN firstly generates paraphrase data via back translation at large scale and then automatically labels the data with syntactic transformations. Based on these labeled data, SCPN trains a neural encoder-decoder model to produce syntactically controlled paraphrased adversarial examples. Since SCPN sometimes greatly changes the meaning of original examples, following the practice of former works \cite{TextFooler,BertAttack}, we add a constraint that the semantic similarity between adversarial examples and original ones should be larger than 0.4.

\subsection{Tasks and Datasets}
Experiments are conducted on two NLP tasks: text classification and natural language inference (NLI). We investigate four datasets which are widely used by existing works of adversarial attack and defense \cite{alzantot,BertAttack,safer,TextFooler}.The detailed statistics of these datasets are shown in Table \ref{tab:dataset}. 
	
\begin{table}
    \centering
    \small
    \setlength{\tabcolsep}{3pt}
    \caption{Dataset statistics.}
    \label{tab:dataset}
    {
        \begin{tabular}{llccc}
        \hline
        \hline
        Task & Dataset & Train & Test & Avg Len\\
        \hline
        \multirow{3}{*}{Classification} & MR & 9K & 1K & 20 \\
        &SST2 & 67K & 1.8K & 20 \\
        &IMDB & 25K & 25K & 215 \\
        \hline
        Entailment & MNLI & 433K & 10K & 11 \\
        \hline
        \hline
        \end{tabular}
    }
\end{table}

\subsubsection{Text Classification}
We use three text classification datasets (phrase-level, sentence-level, document-level) with average text lengths from 20 to 215 words.
 
\begin{itemize}
    \item \textit{SST2} \cite{SST2}: a phrase-level binary sentiment classification dataset on movie reviews;
    \item \textit{MR} \cite{mr}: a sentence-level binary sentiment classification dataset on movie reviews. Following the practice in \cite{TextFooler}, We select 90$\%$ of the data as our training set and 10$\%$ of the data as our test set;
    \item \textit{IMDB} \cite{IMDB} : a document-level binary sentiment classification dataset on movie reviews. 
    
\end{itemize}

\subsubsection{Natural Language Inference}
Natural language inference aims at determining the semantic relationship between a pair of sentences. We use the Multi-Genre Natural Language Inference (MNLI) dataset:
\begin{itemize}
    \item \textit{MNLI} \cite{MNLI}: a widely adopted NLI benchmark with coverage of transcribed speech, popular fiction, and government reports.
\end{itemize}

\subsection{Experimental Setup}
The implementation of anomaly detectors is based on the PyTorch implementation of BERT \footnote{\href{https://github.com/huggingface/pytorch-pretrained-BERT}{https://github.com/huggingface/pytorch-pretrained-BERT}}. We use the base-uncased version of BERT, which has 12 layers, 768 hidden units, 12 heads, and 110M parameters. We use AdamW as our optimizer with a learning rate of 3e-5 and a batch size of 16. The number of epochs is set to 5. The victim model is the base-uncased version of BERT fine-tuned on corresponding datasets. The experiments are conducted with 4 NVIDIA RTX 2080Ti. The hyperparameter $k$ in Section \ref{sec:adaptive} is 5. When attacking natural language inference task, we keep the original premises unchanged and generate adversarial hypotheses.

\section{Experimental Results}
\subsection{Anomaly Detection}
\label{sec:detection}
\subsubsection{Specific Anomaly Detector}

\begin{table*}
	\centering
	\small
	\setlength{\tabcolsep}{3pt}
	\caption{Detection results of specific detector. Acc. is accuracy. Prec. is precision. Rec. is recall. F1. is F1 score. Sentence-level attacks like SCPN are not suitable for the IMDB dataset, given the sentences in IMDB are too long, so their results are unavailable.}
	\label{tab:specific detector}
	{
		\begin{tabular}{l|cccc|cccc|cccc|cccc}
			\hline
			\hline		
			{ } &
        \multicolumn{4}{c|}{MR} &
        \multicolumn{4}{c|}{SST2} &
        \multicolumn{4}{c|}{IMDB} &
        \multicolumn{4}{c}{MNLI} \\
        &
        Acc. &
        Prec. &
        Rec. &
        F1. &
        Acc. &
        Prec. &
        Rec. &
        F1. &
        Acc. &
        Prec. &
        Rec. &
        F1. &
        Acc. &
        Prec. &
        Rec. &
        F1. \\
        \hline
        DeepWordBug & 97.2 & 98.5 & 95.9 & 97.2 & 98.1  & 98.2 & 98.8  & 98.5 & 97.2 & 98.1 & 96.3 & 97.2 & 91.3 & 88.9 & 94.4 & 91.6  \\
        TextFooler & 86.3 & 96.9 & 75.0 & 84.6 & 84.5 & 81.0 & 90.2 & 85.4 & 93.0 & 99.3 & 86.6 & 92.5 & 86.1 & 87.7 & 84.0 & 85.8 \\
        BertAttack & 85.1 & 96.2 & 70.7 & 81.5 & 90.0 & 95.2 & 84.4 & 89.4 & 90.8 & 97.3 & 83.9 & 90.1 & 84.3 & 87.2 & 80.4 & 83.6\\
        SCPN & 93.9 & 94.7 & 93.0 & 93.8 & 91.0 & 88.2  & 94.6 & 91.3 & - & - & - & - & 91.0 & 88.8 & 93.8 & 91.2 \\
			\hline
			\hline
		\end{tabular}
	}

\end{table*}

For each attack method on each dataset, we test the performance of the specific anomaly detector with 500 adversarial examples and 500 normal examples from the test set. We evaluate the performance of the specific anomaly detector with accuracy, precision, recall and F1 score. As shown in Table \ref{tab:specific detector}, for all the attack methods we examined, their adversarial examples can be easily detected with high accuracy. Although contextualized attack methods (e.g. BertAttack) are supposed to be more imperceptible than their non-contextualized counterparts (e.g. TextFooler), our experiment indicates that the performances of BertAttack and TextFooler are comparable in terms of detectability.

\subsubsection{General Anomaly Detector}
During the training stage of the general anomaly detector, the detector has seen the adversarial examples generated by all the four attack methods in the upper experiments of specific anomaly detectors. To verify the performance of the general anomaly detector on unseen attacks, we introduce three new attacks in the test stage: TextBugger \cite{textbugger} (character-level), Alzantot \cite{alzantot} (word-level) and GAN \cite{gan} (sentence-level). The new attacks are marked with * in Table \ref{tab:general detector}. For each attack, we also test the performance of the general detector with 500 normal examples and 500 adversarial examples on each test set.

\begin{table*}
	\centering
	\small
	\setlength{\tabcolsep}{3pt}
	\caption{\label{tab:general detector} Detection results of general detector. Prec. is precision. Rec. is recall. F1. is F1 score.}
	{
        \begin{tabular}{l|cccc|cccc|cccc|cccc}
        \hline
        \hline
        
        { } &
        \multicolumn{4}{c|}{MR} &
        \multicolumn{4}{c|}{SST2} &
        \multicolumn{4}{c|}{IMDB} &
        \multicolumn{4}{c}{MNLI}\\
        &
        Acc. &
        Prec. &
        Rec. &
        F1. &
        Acc. &
        Prec. &
        Rec. &
        F1. &
        Acc. &
        Prec. &
        Rec. &
        F1. &
        Acc. &
        Prec. &
        Rec. &
        F1. \\
        \hline
        DeepWordBug & 98.9 & 99.9 & 98.1 & 98.9 & 97.5 & 96.6 & 98.4 & 97.5 & 96.1 & 98.2 & 93.9 & 96.0 & 92.2 & 88.0 & 97.8 & 92.6  \\
        TextBugger* & 96.0 & 99.5 & 92.4 & 95.9 & 95.6 & 98.6 & 92.4 & 95.4 & 93.2 & 98.3 & 87.9 & 92.8 & 89.1 & 87.2 & 91.6 & 89.4 \\
        TextFooler & 88.8 & 95.1 & 81.8 & 88.0 & 90.4 & 91.1 & 89.6 & 90.3 & 90.1 & 96.5 & 83.2 & 89.4 & 87.3 & 86.8 & 88.0 & 87.4 \\
        BertAttack & 86.2 & 93.7 & 75.3 & 83.5 & 89.4 & 94.5 & 83.6 & 88.7 & 91.6 & 96.2 & 86.7 & 91.2 & 87.1 & 86.7 & 87.6 & 87.2 \\
        Alzantot* & 87.8 & 95.0 & 79.8 & 86.7 & 86.5 & 95.9 & 76.2 & 85.0 & 83.7 & 95.4 & 70.9 & 81.3 & 86.4 & 86.5 & 86.2 & 86.4 \\
        SCPN & 95.0 & 95.7 & 94.3 & 95.0 & 95.0 & 92.2 & 98.3 & 95.1 & - & - & - & - & 90.0 & 87.2 & 93.6 & 90.0\\
        GAN* & 89.9 & 95.2 & 83.9 & 89.2 & 89.2 & 91.2 & 86.7 & 88.9 & - & - & - & - & 85.4 & 86.3 & 84.2 & 85.2\\
        \hline
        \hline
        
        \end{tabular}
    }
    
\end{table*}

As shown in Table \ref{tab:general detector}, our general detector performs consistently well for all attacks, regardless of whether their adversarial examples have been included during the training stage or not. Moreover, in most scenarios, the general detector outperforms the specific one. This result indicates an important finding: for all current unnatural adversarial examples, despite their differences in attack ways, they share some common anomaly features that can be learned by our general anomaly detector. By quantifying these anomaly features and setting corresponding constraint, we will be able to help existing attacks generate more natural and imperceptible adversarial examples.

\subsubsection{Quantifying Anomaly}
We propose the new metric - degree of anomaly to quantify the anomaly of a sentence. As demonstrated in Table \ref{tab:unnatural}, degree of anomaly is able to filter unnatural adversarial examples which cannot be distinguished by other automatic perturbation constraints such as semantic similarity, perturbation rate and increased number of grammar errors. 

\begin{table*}
	\centering
	\small
	\setlength{\tabcolsep}{3pt}
	\caption{\label{tab:unnatural} Examples of the unnatural adversarial sentences that can be filtered by degree of anomaly but not by other perturbation constraints. The threshold for each constraint used in attack is written underneath the data. }
	{
        \begin{tabular}{l|l|c|c|c|c}
        \hline
        \hline
        Original Sentence & Adversarial Sentence & Sem. Sim.& Pert. R. & Gram. Err. & Deg. of Anom.\\
        \hline
        \makecell[l]{\textcolor{blue}{So aggressively} cheery that pollyana\\would reach for a barf bag.} & \makecell[l]{\textcolor{red}{Thereby powerfully} cheery that\\ pollyana would reach for a barf bag.} &  
        \makecell[c]{86.6\%\\\textcolor{green}{(\textgreater40\%)}} & \makecell[c]{18.2\%\\\textcolor{green}{(\textless20\%)}} & \makecell[c]{0\\\textcolor{green}{(=0)}} & \makecell[c]{92.9\%\\\textcolor{red}{(\textgreater50\%)}} \\
        \hline
        \makecell[l]{Raimi and his team couldn't have\\ done any \textcolor{blue}{better} in bringing the story\\ of spider man to the big screen.} & \makecell[l]{Raimi and his team couldn't have\\ done any \textcolor{red}{best} in bringing the story\\ of spider man to the big screen.} &  
        \makecell[c]{85.2\%\\\textcolor{green}{(\textgreater40\%)}} & \makecell[c]{4.8\%\\\textcolor{green}{(\textless20\%)}} & \makecell[c]{0\\\textcolor{green}{(=0)}} & \makecell[c]{97.6\%\\\textcolor{red}{(\textgreater50\%)}} \\
        \hline
        \hline
        
        \end{tabular}
    }
    
\end{table*}

\subsubsection{Attack under anomaly constraint}

\begin{table*}
	\centering
	\small
	\setlength{\tabcolsep}{3pt}
	\caption{\label{tab:constraint} The attack success rate of attacks using BERT as victim model without and with the new constraint on MR, SST2, IMDB, MNLI.}
	{
        \begin{tabular}{l|cc|cc|cc|cc}
        \hline
        \hline
        
        {} &
        \multicolumn{2}{c|}{MR} &
        \multicolumn{2}{c|}{SST2} &
        \multicolumn{2}{c}{IMDB} &
        \multicolumn{2}{c}{MNLI} \\
        &
        w/o Cons. &
        w Cons. &
        w/o Cons. &
        w Cons. &
        w/o Cons. &
        w Cons. &
        w/o Cons. &
        w Cons. \\
        \hline
        Deepwordbug & 82.3 & 8.9  & 78.6 & 2.3  & 74.5 & 28.6 & 76.8 & 33.5\\
        TextFooler & 80.4 & 39.8 & 60.9 & 35.9 & 86.6 & 41.9 & 86.5 & 45.4 \\
        BertAttack & 84.7 & 15.4 & 87.3 & 11.7 & 87.4 & 18.2 & 89.8 & 20.5 \\
        \hline
        \hline
        
        \end{tabular}
    }
\end{table*}

\begin{table*}
	\centering
	\small
	\setlength{\tabcolsep}{3pt}
	\caption{\label{tab:prlms} The attack success rate of attacks without and with the new constraint using different PrLMs as victim models on MR.}
	{
        \begin{tabular}{l|cc|cc|cc}
        \hline
        \hline
        
        {} &
        \multicolumn{2}{c|}{BERT} &
        \multicolumn{2}{c|}{RoBERTa} &
        \multicolumn{2}{c}{ELECTRA} \\
        &
        w/o Cons. &
        w Cons. &
        w/o Cons. &
        w Cons. &
        w/o Cons. &
        w Cons. \\
        \hline
        Deepwordbug & 82.3 & 8.9  & 83.7 & 11.4 & 79.5 & 8.0 \\
        TextFooler & 80.4 & 39.8 & 67.2 & 36.9 & 63.6 & 34.6 \\
        BertAttack & 84.7 & 15.4 & 73.5 & 17.6 & 70.6 & 14.4 \\
        \hline
        \hline
        
        \end{tabular}
    }
\end{table*}

For each attack method on each dataset, we compare the attack success rate with and without the new constraint that the degree of anomaly of $X_{adv}$ should be smaller than 0.5 in Table \ref{tab:constraint}. Since the attack success rate of SCPN is already very low under the constraint of semantic similarity, we do not conduct the experiment with SCPN. The results indicate that under the constraint of degree of anomaly, the attack success rate decreases sharply for all levels of attacks. 

In order to see whether this phenomenon exists for other PrLMs, we further conduct experiments on RoBERTa-base \cite{liu2019roberta} and ELECTRA-base \cite{clark2020electra}. As shown in Table \ref{tab:prlms}, we can also observe a sharp decrease of adversarial success rate under the new constraint for these PrLMs. Such sharp decrease indicates that the PrLMs are more robust than previous attack methods have claimed, given most of the adversarial examples generated by previous attacks are unnatural and detectable. However, there still exist some undetectable adversarial examples that can successfully mislead PrLMs, which pose real potential risks for PrLMs.

\subsection{Randomization}
\label{sec:randomization}

We investigate the effect of four randomization processes \textbf{random synonym substitution}, \textbf{random adverb insertion}, \textbf{random MLM suggestion} and \textbf{back translation} in two settings: (a) randomization after attack and (b) randomization during attack.

In addition, ideal randomization should only invalidate adversarial examples but not change the prediction of the classifier on benign examples. So we investigate whether data augmentation can make the victim model more robust against randomization. For each experiment, we train the PrLM on the original dataset (w/o Aug.) and on its augmented version (w/ Aug.) after the randomization process is applied. We record the prediction accuracy on adversarial examples (after-attack accuracy) in Table \ref{tab:randomization after attack} and Table \ref{tab:randomization during attack} to demonstrate the defense performance. 

\subsubsection{Randomization After Attack}

\begin{table*}
	\centering
	\small
	\setlength{\tabcolsep}{3pt}
	\caption{\label{tab:randomization after attack} The prediction accuracy for randomization after attack, using BERT as victim model and SST2 as dataset. Avg. is the average accuracy of the four attacks. Aug. means data augmentation of randomization strategy during the training of the classifier.}
	{
        \begin{tabular}{l|c|cc|cc|cc|cc}
        \hline
        \hline
        
        {} &
        \multicolumn{1}{c|}{No Rand.} &
        \multicolumn{2}{c|}{Rand. Subs.} &
        \multicolumn{2}{c|}{Rand. Insert.} &
        \multicolumn{2}{c|}{MLM Suggest.} &
        \multicolumn{2}{c}{Back Trans.} \\
        &
        w/o Aug. &
        w/o Aug. &
        w/ Aug. &
        w/o Aug. &
        w/ Aug. &
        w/o Aug. &
        w/ Aug. &
        w/o Aug. &
        w/ Aug.  \\
        \hline
        No Attack & 92.6 & 90.5 & 91.8 & 89.5 & 91.5 & 86.2 & 88.0 & 91.5 & 93.0 \\
        \hdashline
        DeepWordBug & 19.8 & 68.5 & 69.6 & 68.8 & 70.5 & 68.3 & 70.0 & 72.1 &  73.4\\
        TextFooler & 36.2 & 59.6 & 60.4 & 54.9 & 55.6 & 57.2 & 59.4 & 76.6 & 76.3 \\
        BertAttack & 11.8 & 35.2 & 34.5 & 43.7 & 50.6 & 43.1 & 42.4 & 54.1 & 54.5 \\
        SCPN & 71.8 & 81.8 & 82.3 & 82.8 & 83.4 & 82.2 & 82.5 & 81.9 & 82.7   \\
        \hdashline
        Avg. & 34.9 & 61.3 & 61.7 & 62.6 & 65.0 & 62.7 & 63.6 & 71.2 & \textbf{71.7} \\
        \hline
        \hline
        
        \end{tabular}
    }
    
\end{table*}

Table \ref{tab:randomization after attack} shows the performance of different randomization processes applied after the attack. The results indicate that all randomization methods explored are able to enhance the performance of PrLMs against adversarial examples. Among all the randomization methods, back translation is the most effective way to mitigate adversarial effects. Furthermore, using the data augmentation strategy with applied randomization process usually leads to better performance of PrLM.

\subsubsection{Randomization During Attack}
Table \ref{tab:randomization during attack} shows the performance of different randomization processes applied during the dynamic attack. An interesting observation is that, although back translation performs well when applied after the attack, it fails to effectively mitigate adversarial effects in dynamic attack. This phenomenon can be explained by the fact that we leverage another DNN to conduct back translation, and this extra DNN is still vulnerable faced with dynamic attack. For word-level attacks, the performances of all randomization strategies except back translation are better compared with the static scenario. The effectiveness of dynamically applied randomization can be explained in two parts: (1) as experiments of randomization after attack reveal, the randomization process can directly invalidate certain adversarial examples; (2) by forcing the victim model to predict on randomized version of inputs, the dynamic randomization processes manage to perturb the feedback obtained by attack methods and therefore mislead its judgement. For example, word-level attacks need to sort words by their importance scores, which are calculated using prediction scores generated by the victim model. Randomization can perplex these scores and lead to wrong judgement for attack models. 

\begin{table*}
	\centering
	\small
	\setlength{\tabcolsep}{3pt}
	\caption{\label{tab:randomization during attack} The prediction accuracy for randomization during attack, using BERT as victim model and SST2 as dataset. Avg. is the average accuracy of the four attacks. Aug. means data augmentation of randomization strategy during the training of the classifier.}
	{
        \begin{tabular}{l|c|cc|cc|cc|cc}
        \hline
        \hline
        
        {} &
        \multicolumn{1}{c|}{No Rand.} &
        \multicolumn{2}{c|}{Rand. Subs.} &
        \multicolumn{2}{c|}{Rand. Insert.} &
        \multicolumn{2}{c|}{MLM Suggest.} &
        \multicolumn{2}{c}{Back Trans.} \\
        &
        w/o Aug. &
        w/o Aug. &
        w/ Aug. &
        w/o Aug. &
        w/ Aug. &
        w/o Aug. &
        w/ Aug. &
        w/o Aug. &
        w/ Aug.  \\
        \hline
        No Attack & 92.6 & 90.5 & 91.8 & 89.5 & 91.5 & 86.2 & 88.0 & 91.5 & 92.0 \\
        \hdashline
        DeepWordBug & 19.8 & 63.6 & 63.8 & 59.5 & 55.0 & 61.3 & 62.9 & 58.0 & 58.8                  \\
        TextFooler & 36.2 & 74.5 & 77.2 & 71.0 & 52.5 & 67.6 & 68.5 & 36.0 & 38.0 \\
        BertAttack & 11.8 & 68.5 & 69.6 & 61.0 & 38.5 & 59.5 & 60.5 & 8.5 & 9.0\\
        SCPN & 71.8 & 80.8 & 81.8 & 79.5 & 78.0 & 81.0 & 82.1 & 72.8 & 73.0      \\
        \hdashline
        Avg. & 34.9 & 71.9 & \textbf{73.1} & 67.8 & 56 & 67.4 & 68.5 & 43.8 & 44.7 \\
        \hline
        \hline
        \end{tabular}
    }
\end{table*}

\subsection{Universal Defense Framework}
  
\begin{table*}
	\centering
	\small
	\setlength{\tabcolsep}{3pt}
	\caption{\label{tab:defense}The performance of the universal defense framework using BERT as PrLM. Orig\% is the prediction accuracy of normal examples and Adv\% is the prediction accuracy of adversarial examples. The results are based on the average of three runs.}
	{
        \begin{tabular}{l|cc|cc|cc|cc|cc|cc|cc|cc}
        \hline
        \hline
        
        {} &
        \multicolumn{4}{c|}{MR} &
        \multicolumn{4}{c|}{SST2} &
        \multicolumn{4}{c|}{IMDB} &
        \multicolumn{4}{c}{MNLI}\\
        {} &
        \multicolumn{2}{c|}{w/o defense} & \multicolumn{2}{c|}{w/ defense} &
        \multicolumn{2}{c|}{w/o defense} & \multicolumn{2}{c|}{w/ defense} &
        \multicolumn{2}{c|}{w/o defense} & \multicolumn{2}{c|}{w/ defense} &
        \multicolumn{2}{c|}{w/o defense} & \multicolumn{2}{c}{w/ defense} \\
        &
        Orig\%&
        Adv\%&
        Orig\%&
        Adv\%&
        Orig\%&
        Adv\%&
        Orig\%&
        Adv\%&
        Orig\%&
        Adv\%&
        Orig\%&
        Adv\%&
        Orig\%&
        Adv\%&
        Orig\%&
        Adv\%\\
        \hline
        Deepwordbug & 86.2 & 15.3 & 86.4 & 56.3 & 92.6 & 19.8 & 92.4 & 61.7 & 92.4 & 23.6 & 92.3 & 80.5 & 84.0 & 18.7 & 82.7 & 69.3 \\
        TextBugger* & 86.2 & 19.7 & 86.4 & 60.1 & 92.6 & 30.6 & 92.4 & 67.7 & 92.4 & 12.5 & 92.3 & 77.4 & 84.0 & 14.4 & 82.7 & 63.7\\
        TextFooler & 86.2 & 16.9 & 86.4 & 66.2 & 92.6 & 36.2 & 92.4 & 72.3 & 92.4 & 12.4 & 92.3 & 89.2 & 84.0 & 11.3 & 82.7 & 68.1 \\
        BertAttack & 86.2 & 13.2 & 86.4 & 60.5 & 92.6 & 11.8 & 92.4 & 65.1 & 92.4 & 11.6 & 92.3 & 85.4 & 84.0 & 9.5 & 82.7 & 66.5\\
        Alzantot* &  86.2& 22.3 & 86.4 &  68.4 & 92.6 & 43.2& 92.4& 73.2& 92.4& 14.1& 92.3& 82.7& 84.0& 20.2& 82.7& 64.9\\
        SCPN & 86.2 & 67.2 & 86.4 & 73.9 & 92.6 & 71.8 & 92.4 & 81.8 & 92.4 & - & 92.3 & - & 84.0 & 66.9 & 82.7 & 75.1 \\
        GAN* & 86.2 & 72.9 & 86.4 & 79.0 & 92.6 & 79.1 & 92.4 & 85.4 & 92.4 & - & 92.3 & - & 84.0 & 70.6 & 82.7 & 77.6 \\
        \hline
        \hline
        
        \end{tabular}
    }
    
\end{table*}

\begin{table*}
	\centering
	\small
	\setlength{\tabcolsep}{3pt}
	\caption{\label{tab:comparison} The performance of our universal defense framework compared with other word-level defenses using BERT as PrLM. Orig\% is the prediction accuracy of normal examples and Adv\% is the prediction accuracy of adversarial examples. The results are based on the average of three runs.}
	{
        \begin{tabular}{l|cc|cc|cc|cc}
        \hline
        \hline
        
        {} &
        \multicolumn{2}{c|}{MR} &
        \multicolumn{2}{c|}{SST2} &
        \multicolumn{2}{c|}{IMDB} &
        \multicolumn{2}{c}{MNLI} \\
        &
        Orig\%&
        Adv\%&
        Orig\%&
        Adv\%&
        Orig\%&
        Adv\%&
        Orig\%&
        Adv\% \\
        \hline
        No Defense & 86.2 & 16.9 & 92.6 & 36.2 & 92.4 & 12.4 & 84.0 & 11.3\\
        \hdashline
        Adv Training & 85.6 & 34.6 & 92.1 & 48.2 & 92.2 & 34.2 & 82.3 & 33.4\\
        DISP & 82.0 & 42.2 & 91.1 & 69.9 & 91.7 & 82.0 & 76.3 &  35.1\\
        SAFER & 79.0 & 55.4 & 90.8 & \textbf{75.1} & 91.3 & 88.1 & 82.1  & 54.7\\
        Ours & \textbf{86.4} & \textbf{66.2} & \textbf{92.4} & 72.3 & \textbf{92.3} & \textbf{89.2} & \textbf{82.7} & \textbf{68.1}  \\
        \hline
        \hline
        \end{tabular}
    }
\end{table*}

Based on the results of Table \ref{tab:randomization during attack}, we adopt random substitution as the randomization process in the universal defense framework. Table \ref{tab:defense} shows the performance of BERT with and without the universal defense framework when faced with different attacks. To verify the universality of the defense framework, we also conduct experiments on three unseen attacks which are the same attacks in Table \ref{tab:general detector}. The unseen attacks are marked with * in Table \ref{tab:defense}. Since randomization may lead to little variance of results, we report the results based on the average of three runs. Experimental results indicate that the proposed defense framework can effectively help PrLM defend against textual adversarial attack. Furthermore, thanks to the anomaly detection, the randomization process is only applied to detected adversarial examples. For the very few benign sentences that are detected by mistake as anomaly and then randomized, the data augmentation in the training stage ensures that the classifier can make correct predictions on most randomized benign sentences. Therefore the proposed framework does not hurt the prediction accuracy for non-adversarial examples, which is important in real application scenarios.

We compare the performance of our universal defense framework with several state-of-the-art word-level defenses (adversarial training, DISP, SAFER) while facing TextFooler as the attack model. DISP \cite{DISP} detects and restores adversarial examples by leveraging a perturbation discriminator and an embedding estimator. SAFER \cite{safer} smooths the classifier by averaging the outputs of a set of randomized examples. As shown in Table \ref{tab:comparison}, although DISP and SAFER are especially designed for word-level attacks, our universal defense framework outperforms them in most cases on both original accuracy and after-attack accuracy. 

\section{Analysis}
\subsection{Impact of Number of Adversarial Examples}

\begin{figure*}
  \centering
  \includegraphics[width=0.88\textwidth]{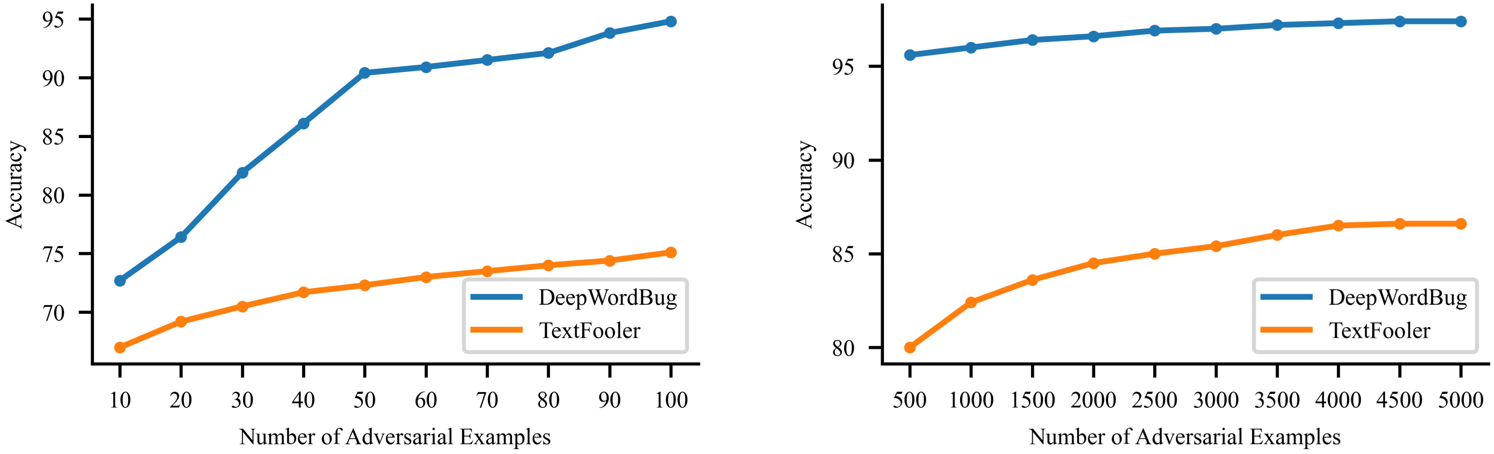}
  \centering
  \caption{\label{fig:deepwordbug}Accuracy of the specific detector against DeepWordBug and against TextFooler trained with different numbers of adversarial examples.}
\end{figure*}

We conduct experiments to investigate how the performance of the detector changes with the number of adversarial examples used in training. We use $n$ adversarial examples and $n$ normal examples to train the anomaly detector and observe the accuracy of the anomaly detector with different $n$. 

Figure \ref{fig:deepwordbug} shows the accuracy of the anomaly detector against DeepWordBug and TextFooler with different numbers of adversarial examples. We can observe that the accuracy of the detector against DeepWordBug increases very fast from 72.7\% to 95.0\% within 100 adversarial examples, but the accuracy of the detector against TextFooler increases slowly from 67.0\% to 74.6\% within 100 adversarial examples. When the number of adversarial examples $n$ continues to increase, the increase of the accuracy of the detector against DeepWordBug is slow and limited. But the accuracy of the detector against TextFooler keeps increasing markedly until $n$ reaches 4000.

From these results, we could infer that the detector learns very fast to distinguish adversarial examples with character-level perturbations. A limited number of such adversarial examples is enough to reach a satisfying accuracy. But it is more difficult and slow for the detector to learn to distinguish adversarial examples with word-level perturbations. A probable explanation of this difference is that character-level attacks generate many out-of-vocabulary (OOV) words, which is obvious and shallow for the detector to learn, but the anomaly information of word-level adversarial examples is in semantics or grammar, which is a deeper anomaly for the detector to learn.

\subsection{Distribution of Degree of Anomaly}
\begin{figure}
  \centering
  \includegraphics[width=8cm]{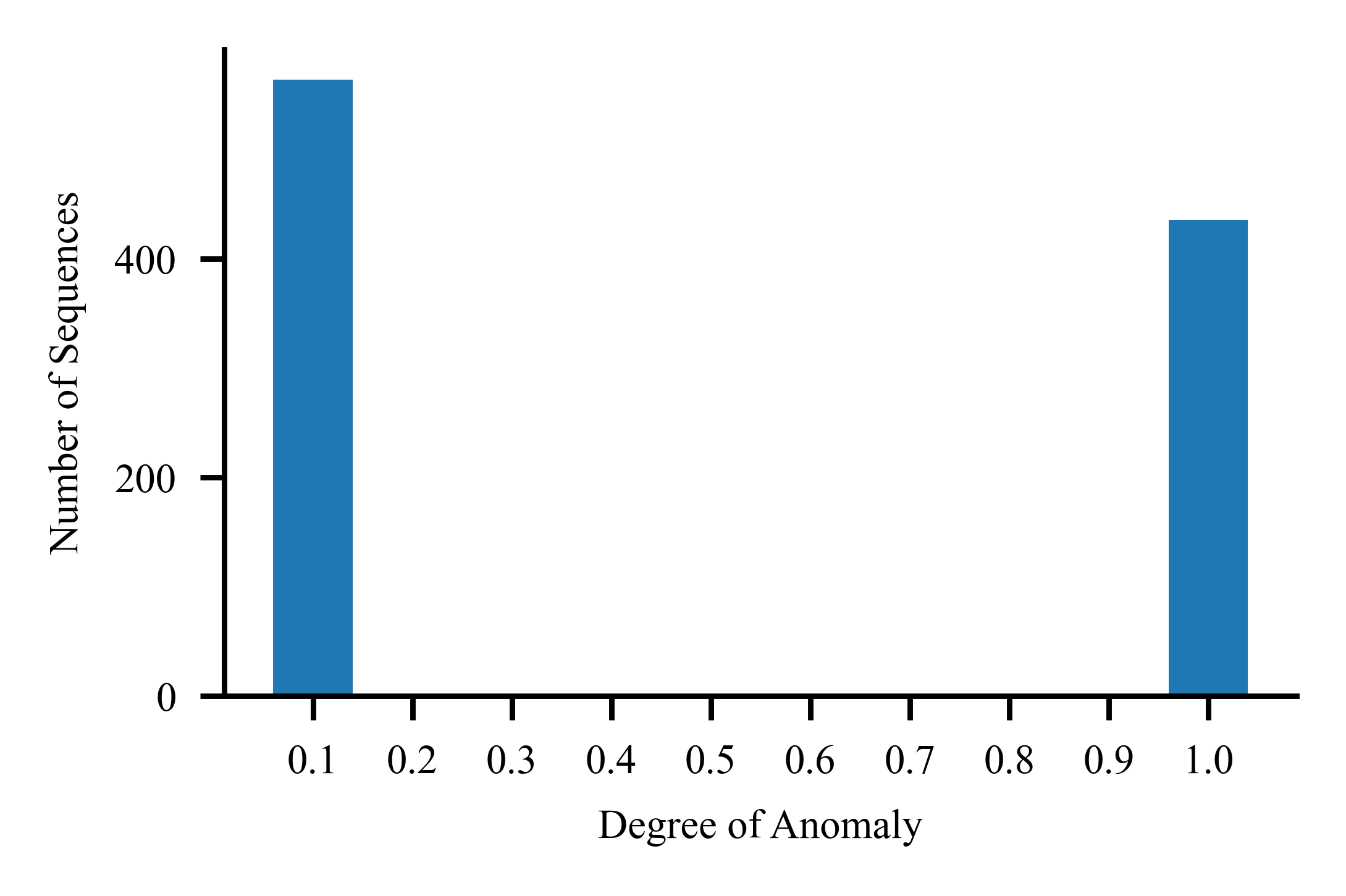}\
  \centering
  \caption{\label{fig:distribution}Distribution of degree of anomaly of 1000 sequences composed of 500 adversarial examples and 500 normal examples, where the x-axis represents the degree of anomaly.}
\end{figure}

We set the threshold of degree of anomaly to be 0.5 because the distribution of degree of anomaly is very polarized. We plot the distribution of the degree of anomaly of 1000 sequences composed of 500 adversarial examples and 500 normal examples in Figure \ref{fig:distribution}. For an input sequence, the probability that it is an adversarial example predicted by the general anomaly detector is either between 0 and 0.1 or between 0.9 and 1.0. 

\subsection{Visualization of Detectable and Undetectable Adversarial Examples}
We adopt a visualization approach to verify that adversarial examples share some common anomaly features which can be learned by the general anomaly detector to distinguish them from normal examples. We extract the last hidden-layer representation $h_0 \in \mathbb{R} ^H$ of the special token \texttt{[CLS]} learned by the general anomaly detector and apply t-SNE\cite{tsne} to plot their latent space representations in 2D. We use the special token \texttt{[CLS]} to get the representation because it is the sentence-level representation containing the information of the whole sentence and is especially designed for sentence classification. Since we use the \texttt{[CLS]} token in the training and inference of the detector, we also use \texttt{[CLS]} to plot the representation for consistency. 

\begin{figure}
  \centering
  \includegraphics[width=6cm]{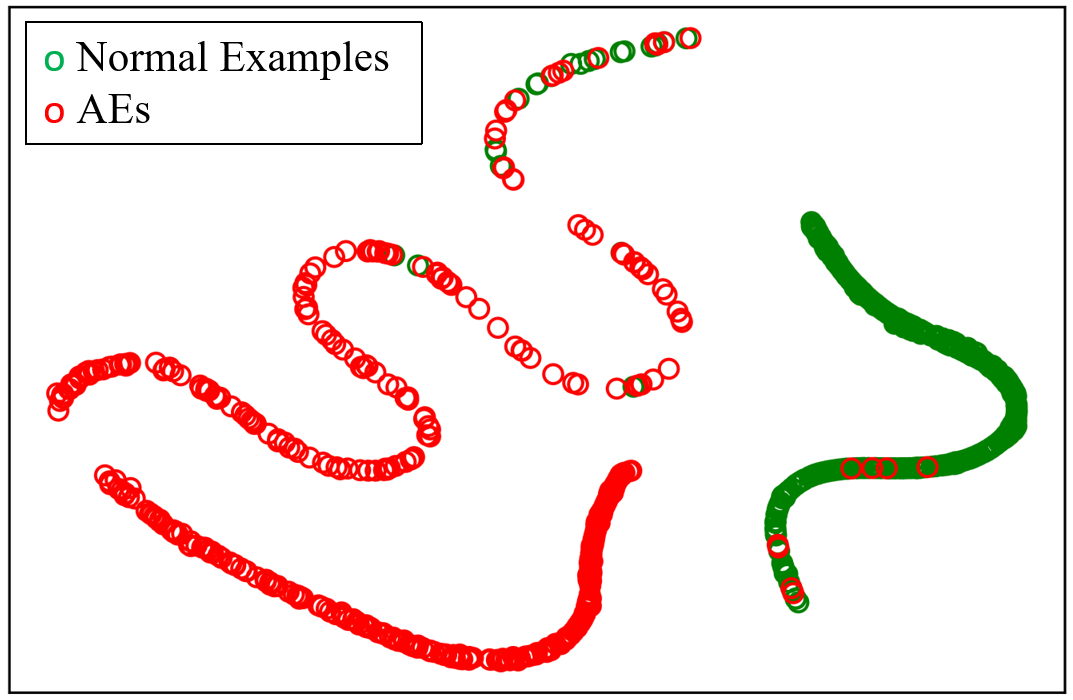}
  \centering
  \caption{\label{fig:tsne}Latent space visualization of normal examples and adversarial examples.}
\end{figure}

We randomly select 400 normal examples and 400 adversarial examples (100 adversarial examples for each of the four attack methods) in the SST2 test set. As shown in Figure \ref{fig:tsne}, the latent space representations of adversarial examples and normal examples are well separated, and adversarial examples generated by different attacks share similar representations.

\begin{figure}
  \centering
  \includegraphics[width=6cm]{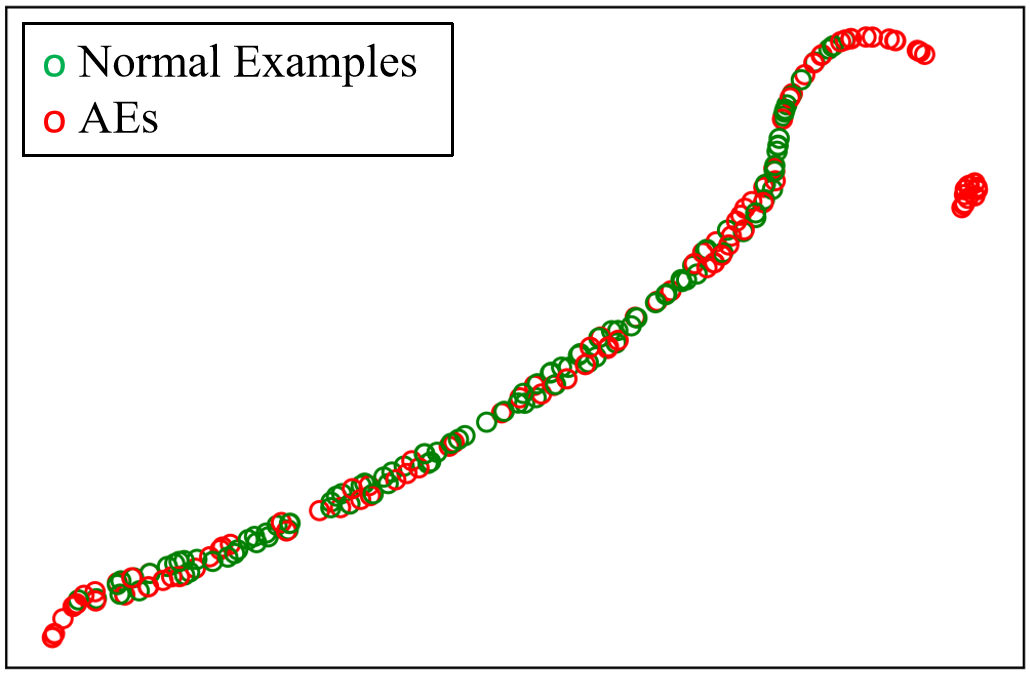}\
  \centering
  \caption{\label{fig:tsne2}Latent space visualization of normal examples and undetectable adversarial examples.}
\end{figure}

To verify that undetectable adversarial examples are indeed more natural than detectable adversarial examples, we apply t-SNE to plot the latent space representations of these undetectable adversarial examples. We randomly select 200 undetectable adversarial examples of TextFooler and 200 normal examples in the SST2 dataset. As shown in Figure \ref{fig:tsne2}, the latent space representations of the normal examples are closely surrounded by the undetectable adversarial examples.

\subsection{Adaptive Attack to Bypass the Randomization} \label{sec:adaptive}
In CV, an adaptive attack called Expectation Over Transformations (EOT) \cite{FalseSecurity} \cite{EOT} is widely used to bypass the defenses based on random transformations (e.g., random combination of input transformations \cite{cv-rand3},  set of random input transformations \cite{cv-rand4}). For a defense model $f()$ that implements a random transformation $t()$ on input $x$, where $t()$ is sampled from a transformation distribution $T$, EOT optimized the expectation over the transformations: $\mathbb{E}_{t \sim T} f(t(x))$. In CV, this optimization problem can be solved using gradient descent because the image input is continuous. Noting that $\nabla \mathbb{E}_{t \sim T} f(t(x)) = \mathbb{E}_{t \sim T} \nabla f(t(x))$, EOT differentiates through the classifier and transformation, and approximates the expectation at each gradient descent step. Borrowing the idea of EOT in CV, we design adaptive attacks trying to bypass the randomization. However, we cannot use the gradient descent as optimization method since textual input is discrete. For the original attack, in each iteration, the attack generates an intermediate adversarial example $x_{int}$ and sends it to the victim model to get its score $f(x_{int})$. For adaptive attack, in each iteration, the attack generates $x_{int}$ and $k$ randomized intermediate adversarial examples $t_i(x_{int})$, $i \in \{1, ..., k\}$, and sends these $k$ sentences to the victim model to get $k$ scores $f(t_i(x_{int}))$. The final score is the expectation over randomizations: $\mathbb{E}_{i \in \{1, ..., k\}}f(t_i(x_{int}))$. 

\begin{table}
    \centering
    \small
    \setlength{\tabcolsep}{3pt}
    \caption{The after-attack accuracy of original attack and adaptive attack for TextFooler and BertAttack when adopting randomization as a defense module.}
    \label{tab:adaptive attack}
    {
        \begin{tabular}{l|cc|cc}
        \hline
        \hline
        {} & \multicolumn{2}{c|}{TextFooler} &
        \multicolumn{2}{c}{BertAttack}
        \\
        {} & Orig. Att. & Adapt. Att. & Orig. Att. & Adapt. Att. \\
        \hline
        MR & 66.2 & 60.1 & 60.5 & 53.1 \\
        SST2 & 72.3 & 65.8 & 65.1 & 59.3\\
        IMDB & 89.2 & 87.3 & 85.4 & 82.6\\
        MNLI & 68.1 & 63.4 & 66.5 & 60.2\\
        \hline
        \hline
        \end{tabular}
    }
\end{table}

We conduct experiments of adaptive attacks based on TextFooler and BertAttack on four datasets when using random synonym substitution as defense strategy. As shown in Table \ref{tab:adaptive attack}, the decrease of after-attack accuracy brought by adaptive attack is quite limited. We explain why EOT in textual attacks fail to perform as good as in image attacks by analysing the differences between textual and image attacks:

\begin{itemize}

    \item \textit{Search space.} Since images are composed of continuous pixels, the search space for adversarial images is infinite and continuous. But for text, the search space of meaningful adversarial sentences is finite and discrete. For instance, in word-level attacks which substitute several words by their synonyms, the search space can be seen as a finite combination of synonyms of all words.

    \item \textit{Optimization process.} The aim of image and text attacks are the same optimization problem: finding the less perturbed adversarial example that can alter the prediction of the model. But their processes to find the optimal solution are quite different. Most image attacks solve the optimization problem via gradient descent, while most textual attacks do not depend on gradients because of the incontinuous search space. As explained in \cite{combinatorial}, word-level attacks can be seen as a combinatorial optimization problem. The optimization process is iteratively trying different combinations and querying the model for feedback decisions. This difference in optimization processes leads to different roles of randomization in resisting image and textual attacks.
    
    \item \textit{Role of randomization.} As most image attacks are gradient-based, the role of randomization is to cause stochastic gradients\cite{FalseSecurity}. But most textual attacks are score-based or decision-based attacks. The role of randomization is to perplex the feedback score or decision received by the attacker. EOT is effective in image attacks because it can estimate the gradient of the stochastic function in order to find the correct descent direction towards the optimal solution. But this is not the case in the gradient-irrelevant optimization process of textual attacks. The utility of EOT in textual attacks is limited to filtering adversarial examples that are unstable to randomizations. However, as the search space of textual adversarial examples is finite and the optimization process is to iteratively query the victim model, there may be too few eligible robust adversarial sentences to be found in limited query times. 
    
\end{itemize} 

\section{Discussion} \label{sec:discussion}
In this section, we discuss the limitation of our work and some directions that we will study in the future.

A limitation of our work is that the general anomaly detector is trained with the samples generated by several specific attack algorithms, which may limit its universality. However, this limitation is not so obvious when we are considering adversarial examples. Because for the attacks of the same granularity, although the attack methods are different in their specific realizations, the generated adversarial examples have similar forms. For example, the character-level attacks always generate sentences with typos and the word-level attacks always generate adversarial samples that replace a portion of words with their synonyms. But it is true that there exist some distribution differences in adversarial samples generated by different attack algorithms, which limit the performance of the general anomaly detector.

There are three directions that we will study in the future as an extension of this work.

1) \textit{Towards stronger detector.}
The anomaly detector in this work is a simple but effective one. To further enhance its performance, there are several possible directions: (\romannumeral1) adopting contrastive learning strategy, which has been proved to be effective in out-of-distribution detection in CV \cite{csi}; (\romannumeral2) using the undetectable adversarial examples generated by attacks under anomaly constraint to further train the detector, and repeating the process iteratively, towards a stronger detector; (\romannumeral3) using the property that randomization on adversarial examples often alter the prediction of model as a complementary measure to help the judgement of detector. 

2) \textit{Rethinking perturbation constraints.}
Existing textual adversarial attacks have achieved a very high attack success rate, but the quality of generated adversarial examples is not satisfying. Many adversarial examples can be distinguished by humans and also by machine. The existence of these unqualified adversarial examples indicates that existing metrics to constrain perturbations are far from perfect. However, existing works of textual attacks \cite{alzantot, deepwordbug, TextFooler} mostly focus on achieving higher attack success rates, but pay little attention to the study of perturbation constraints. We call for more works along the direction of quantifying anomaly or unnaturalness, in order to generate really natural and undetectable adversarial examples.

3) \textit{More than adversarial attacks.}
In fact, we can view adversarial sentences as a kind of out-of-distribution data, which are the noises out of the normal distribution of benign data. There may be a relation between the degree of anomaly and quantifying data uncertainty in text. We will investigate this in the future.

\section{Conclusion\label{sec:conclusion}}
This paper reviews the essence of textual adversarial attacks by revealing four facts: (\romannumeral1) Most existing textual adversarial examples of all levels are unnatural and can be easily detected by machine.  (\romannumeral2) Adversarial examples share some common features and can be quantified by our proposed metric: Degree of Anomaly, which can help future textual attacks generate more natural adversarial examples to evaluate the robustness of PrLMs. (\romannumeral3) Several randomization processes can effectively defend existing textual attacks. (\romannumeral4) Future textual adversarial attacks should focus more on how to overcome detection and resist randomization, otherwise they would be defended by our proposed universal defense framework. In conclusion, our work lays a foundation for future studies of textual attacks and defenses in terms of both analysis tools and practical applications.

\bibliographystyle{IEEEtran}

\bibliography{main}

%
\vspace{-12mm}
\begin{IEEEbiography}[{\includegraphics[width=1in,height=1.25in,clip,keepaspectratio]{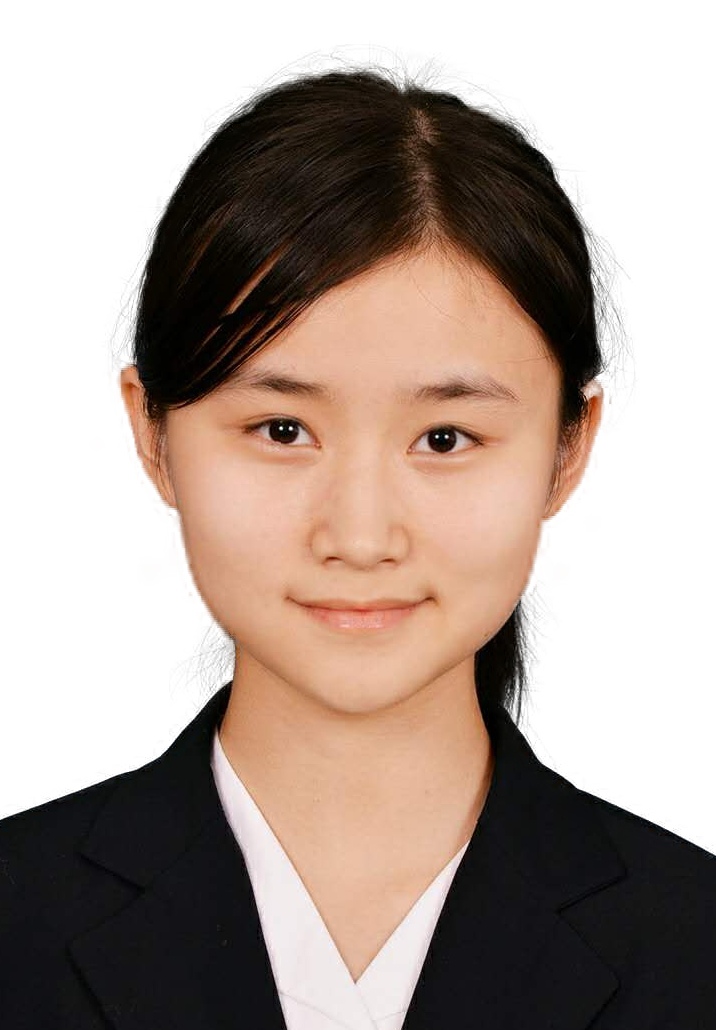}}]{Jiayi Wang}
	received her Bachelor's degree in information engineering from Shanghai Jiao Tong University in 2020. She is working towards her M.S. degree in information engineering with the Center for Brain-like Computing and Machine Intelligence of Shanghai Jiao Tong University and is doing an internship at Ant Group. Her research interests include natural language processing, adversarial attacks and machine reading comprehension. 
\end{IEEEbiography}

\vspace{-12mm}

\begin{IEEEbiography}[{\includegraphics[width=1in,height=1.25in,clip,keepaspectratio]{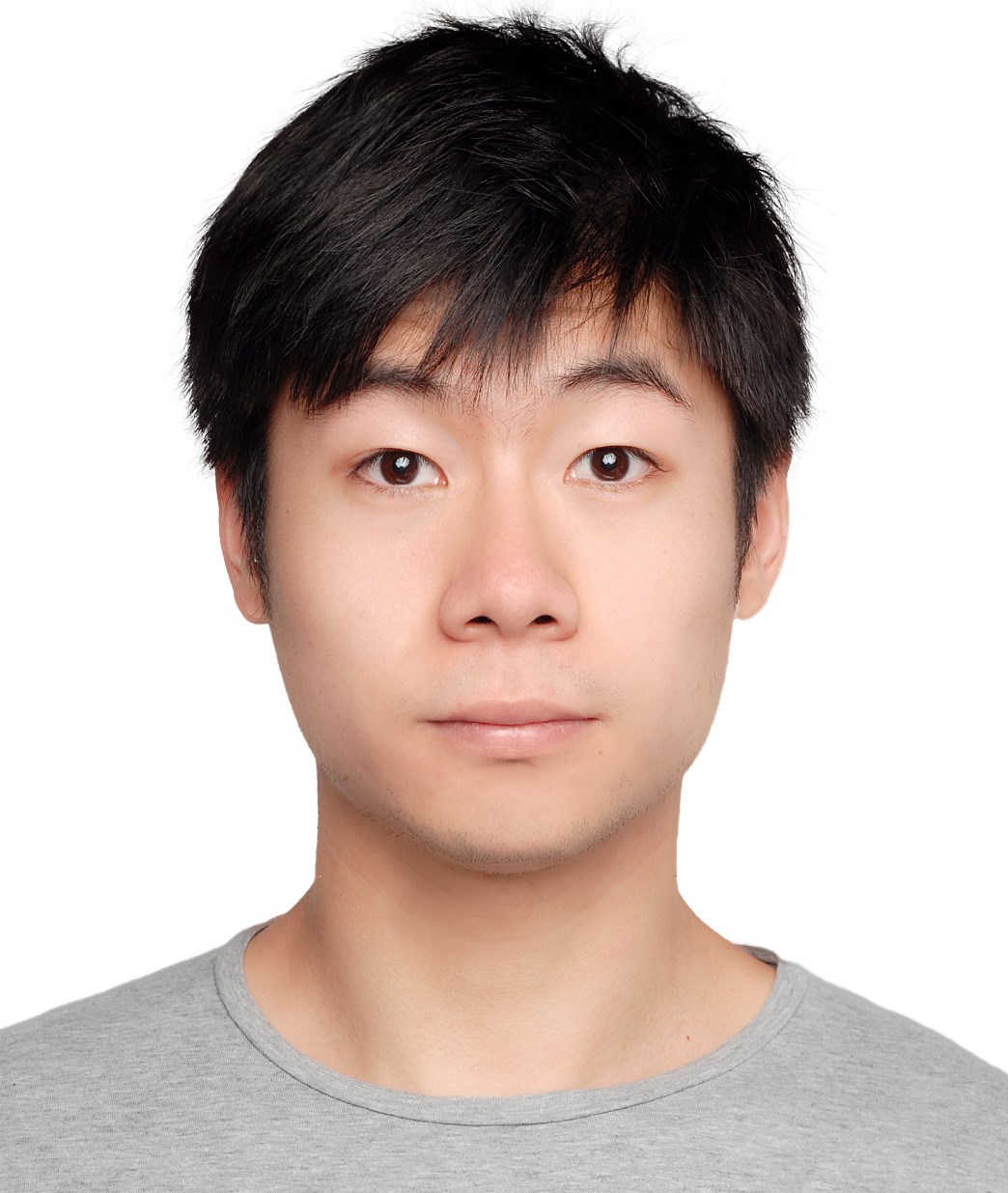}}]{Rongzhou Bao} received his Bachelor's degree in information engineering from Shanghai Jiao Tong University in 2019, his M.S. degree in information engineering from Shanghai Jiao Tong University in 2022. He is currently a machine learning engineer in Ant Group. His research interests include natural language processing, adversarial attacks and pre-trained language models.
\end{IEEEbiography}

\vspace{-12mm}

\begin{IEEEbiography}[{\includegraphics[width=1in,height=1.25in,clip,keepaspectratio]{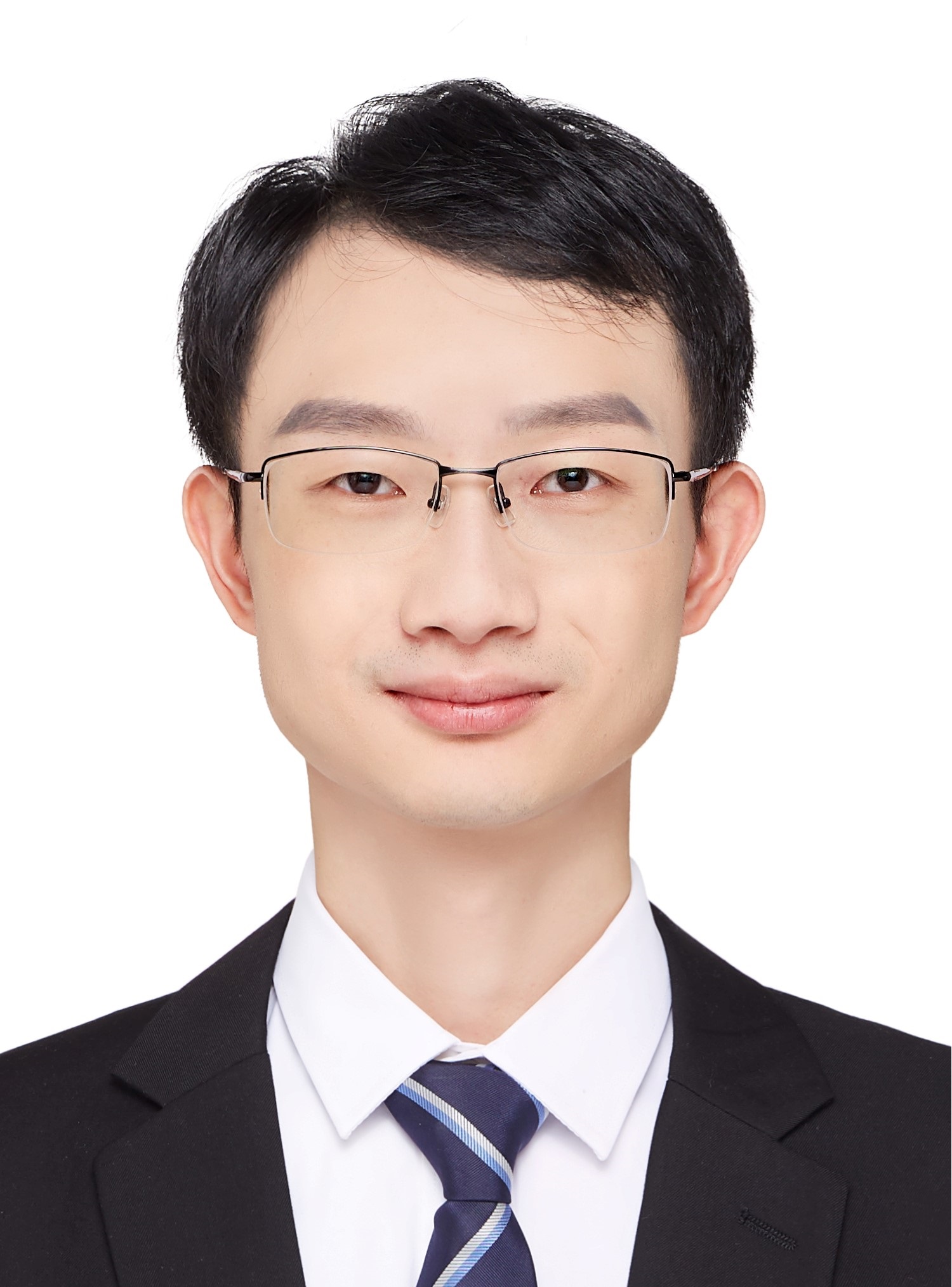}}]{Zhuosheng Zhang} received his Bachelor's degree in internet of things from Wuhan University in 2016, his M.S. degree in computer science from Shanghai Jiao Tong University in 2020. He is working towards his Ph.D. degree in computer science with the Center for Brain-like Computing and Machine Intelligence of Shanghai Jiao Tong University. He was an internship research fellow at NICT from 2019-2020. His research interests include natural language processing, machine reading comprehension, dialogue systems, and machine translation. 
\end{IEEEbiography}

\vspace{-12mm}

\begin{IEEEbiography}[{\includegraphics[width=1in,height=1.25in,clip,keepaspectratio]{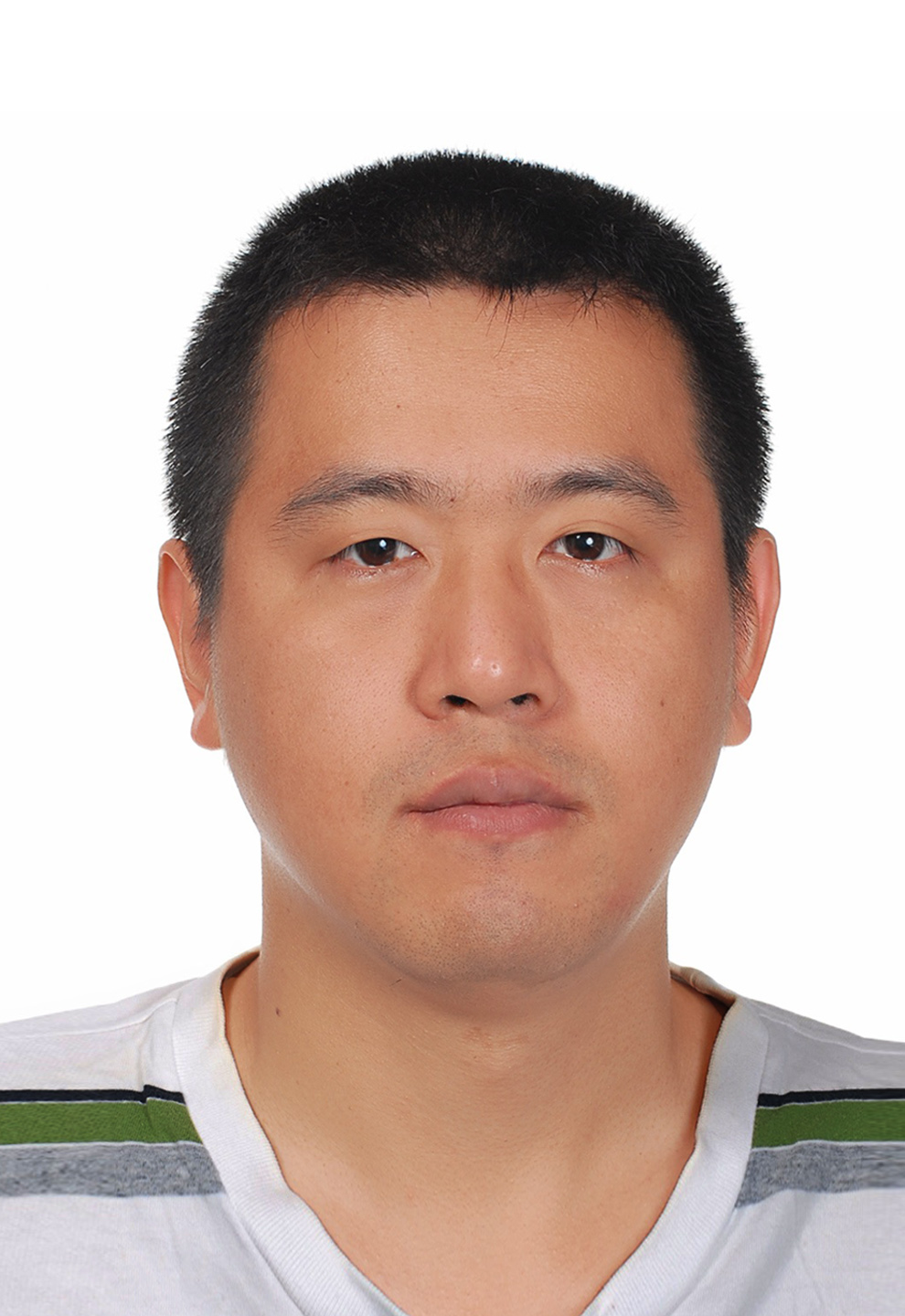}}]{Hai Zhao}
received the BEng degree in sensor and instrument engineering, and the MPhil degree in control theory and engineering from Yanshan University in 1999 and 2000, respectively,
	and the PhD degree in computer science from Shanghai Jiao Tong University, China in 2005. 
	He is currently a full professor at department of computer science and engineering,  Shanghai Jiao Tong University after he joined the university in 2009. 
	He was a research fellow at the City University of Hong Kong from 2006 to 2009, a visiting scholar in Microsoft Research Asia in 2011, a visiting expert in NICT, Japan in 2012.
	He is an ACM professional member, and served as area co-chair in ACL 2017 on Tagging, Chunking, Syntax and Parsing, (senior) area chairs in ACL 2018, 2019 on Phonology, Morphology and Word Segmentation.
	His research interests include natural language processing and related machine learning, data mining and artificial intelligence.
\end{IEEEbiography}

\end{document}